\documentclass[times,twocolumn,final]{elsarticle}
\usepackage[table]{xcolor}
\usepackage{cag}
\usepackage{amssymb}
\usepackage{graphicx}
\usepackage{arydshln} 
\usepackage{multirow}
\usepackage{booktabs}
\usepackage{subcaption}
\usepackage{hyperref}
\hypersetup{
    colorlinks=true,
    linkcolor=blue,
    filecolor=magenta,      
    urlcolor=cyan,
}

\usepackage{amsmath}
\usepackage{amsfonts}
\usepackage{orcidlink}

\usepackage{colortbl}
\definecolor{lightgray}{gray}{0.9}

\newcommand{\highlight}[1]{\textcolor{black}{#1}}
\newcommand{\highlightx}[1]{\textcolor{black}{#1}}

\def\onedot{.}
\def\eg{\emph{e.g}\onedot} 
\def\ie{\emph{i.e}\onedot} 
\def\cf{\emph{cf}\onedot}

\def\etal{\emph{et al}\onedot}

\journal{Computers \& Graphics}

\begin{document}

\begin{frontmatter}

\title{SketchANIMAR: Sketch-Based 3D Animal Fine-Grained Retrieval}

\author[1,3]{Trung-Nghia Le\orcidlink{0000-0002-7363-2610}} 
\author[4]{Tam V. Nguyen\orcidlink{0000-0003-0236-7992}}
\author[1,3]{Minh-Quan Le\orcidlink{0000-0002-0709-3225}}
\author[1,3]{Trong-Thuan Nguyen\orcidlink{0000-0001-7729-2927}}
\author[1,3]{Viet-Tham Huynh\orcidlink{0000-0002-8537-1331}}
\author[1,3]{Trong-Le Do\orcidlink{0000-0002-2906-0360}}
\author[1,3]{Khanh-Duy Le\orcidlink{0000-0002-8297-5666}}
\author[1,3]{Mai-Khiem Tran\orcidlink{0000-0001-5460-0229}}
\author[1,3]{Nhat Hoang-Xuan\orcidlink{0000-0001-8759-0354}}
\author[1,3]{Thang-Long Nguyen-Ho\orcidlink{0000-0003-1953-7679}}
\author[2,3]{Vinh-Tiep Nguyen\orcidlink{0000-0003-4260-7874}}
\author[1,3]{Nhat-Quynh Le-Pham}
\author[1,3]{Huu-Phuc Pham}
\author[1,3]{Trong-Vu Hoang}
\author[1,3]{Quang-Binh Nguyen}
\author[1,3]{Trong-Hieu Nguyen-Mau\orcidlink{0000-0003-2823-3861}}
\author[1,3]{Tuan-Luc Huynh\orcidlink{0000-0002-2481-0724}}
\author[1,3]{Thanh-Danh Le}
\author[1,3]{Ngoc-Linh Nguyen-Ha}
\author[1,3]{Tuong-Vy Truong-Thuy}
\author[1,3]{Truong Hoai Phong}
\author[1,3]{Tuong-Nghiem Diep}
\author[1,3]{Khanh-Duy Ho}
\author[1,3]{Xuan-Hieu Nguyen}
\author[1,3]{Thien-Phuc Tran}
\author[1,3]{Tuan-Anh Yang}
\author[1,3]{Kim-Phat Tran}
\author[1,3]{Nhu-Vinh Hoang}
\author[1,3]{Minh-Quang Nguyen}
\author[1,3]{Hoai-Danh Vo}
\author[1,3]{Minh-Hoa Doan}
\author[1,3]{Hai-Dang Nguyen\orcidlink{0000-0003-0888-8908}}
\author[5]{Akihiro Sugimoto\orcidlink{0000-0001-9148-9822}}
\author[1,3]{Minh-Triet Tran\orcidlink{0000-0003-3046-3041}\corref{cor1}} \ead{tmtriet@fit.hcmus.edu.vn}

\affiliation[1]{organization={University of Science, VNU-HCM}, city={Ho Chi Minh City}, country={Vietnam}}
\affiliation[3]{organization={Vietnam National University}, city={Ho Chi Minh City}, country={Vietnam}}
\affiliation[4]{organization={University of Dayton}, city={Ohio}, country={U.S.}}
\affiliation[2]{organization={University of Information Technology, VNU-HCM}, city={Ho Chi Minh City}, country={Vietnam}}
\affiliation[5]{organization={National Institute of Informatics}, city={Tokyo}, country={Japan}}

\cortext[cor1]{Corresponding author}

\received{\today}

\begin{abstract}

The retrieval of 3D objects has gained significant importance in recent years due to its broad range of applications in computer vision, computer graphics, virtual reality, and augmented reality. However, the retrieval of 3D objects presents significant challenges due to the intricate nature of 3D models, which can vary in shape, size, and texture, and have numerous polygons and vertices. To this end, we introduce a novel SHREC challenge track that focuses on retrieving relevant 3D animal models from a dataset using sketch queries and expedites accessing 3D models through available sketches. Furthermore, a new dataset named ANIMAR was constructed in this study, comprising a collection of 711 unique 3D animal models and 140 corresponding sketch queries. Our contest requires participants to retrieve 3D models based on complex and detailed sketches. We receive satisfactory results from eight teams and 204 runs. Although further improvement is necessary, the proposed task has the potential to incentivize additional research in the domain of 3D object retrieval, potentially yielding benefits for a wide range of applications. We also provide insights into potential areas of future research, such as improving techniques for feature extraction and matching and creating more diverse datasets to evaluate retrieval performance.
\end{abstract}





\end{frontmatter}


\section{Introduction}
\label{sec:introduction}

The rapid development of 3D technologies has produced a remarkable number of 3D objects. Consequently, 3D object retrieval has garnered considerable attention and is beneficial in real-life applications~\cite{stotko2019slamcast,liu2019real, wang2020rgb2hands}, including but not limited to video games, artistic pursuits, cinematography, and virtual reality.

Sketch-based 3D object retrieval aims to retrieve 3D models from a user's hand-drawn 2D sketch. Due to the innate intuitive appeal of freehand drawings, sketch-based 3D object retrieval has drawn a significant amount of attention and is being utilized in numerous critical applications such as 3D scene reconstruction~\cite{guo2022neural, yookwan2022multimodal, li2022high}, 3D geometry video retrieval~\cite{gumeli2022roca, manda2022sketchcleannet, salihu2023sgpcr}, and 3D augmented/virtual reality entertainment~\cite{koca2019augmented, guo2023vid2avatar}. However, sketch-based 3D object retrieval poses a formidable challenge in 3D object retrieval research, primarily due to the large discrepancy between the 2D and 3D modalities: non-realistic 2D sketches differ significantly from their 3D counterparts and respective views.

Several SHREC challenge tracks~\cite{Li-SHREC2012, Li-SHREC2013, Li-SHREC2014, Juefei-SHREC2018, Juefei-SHREC2019, Qin-SHREC2022} have been organized to facilitate research on sketch-based 3D object retrieval. However, the existing datasets incorporated in these tracks primarily comprise generic objects with simplistic shapes and poses. To augment sketch-based 3D object retrieval research, we organize a new SHREC challenge track dedicated to \textit{\textbf{Sketch}-based 3D \textbf{ANIMA}l model fine-grained \textbf{R}etrieval (\textbf{SketchANIMAR})}\footnote{\url{https://aichallenge.hcmus.edu.vn/sketchanimar}}. This track aims to retrieve relevant 3D animal models from a dataset using sketch queries and expedites accessing 3D models through available sketches. Previous SHREC challenge tracks have focused on a limited number of general object categories, often lacking realism. Our challenge track for SHREC 2023 is significantly more challenging and can simulate real-life scenarios more effectively than its predecessors. \highlight{After the challenge concluded, the dataset has been made publicly available for academic purposes.}

First, conventional 3D object retrieval tasks consider only the object category, where the training and test samples are characterized by the same category settings. Consequently, features extracted from these methods are often optimized to fit the seen categories while lacking generalizability for unseen categories. Under such circumstances, the classification-based retrieval embedding learning methods become invalid in practice. Meanwhile, open-set 3D object retrieval can address this issue more effectively by dealing with unseen categories better. This technique involves training retrieval and representation models using seen-category 3D objects, with unseen-category 3D data subsequently used for retrieval. Nevertheless, our fine-grained retrieval task requires participants to conduct an accurate search to get 3D animal models whose shapes correspond to the query, necessitating consideration of unseen categories and poses (\cf~Table~\ref{tab:3d_object_retrieval_tasks}). Compared to searching for 3D general objects of a given category, 3D animal model fine-grained retrieval poses a more significant challenge due to the substantial discrepancy in animal breeds and poses. 

\begin{figure*}[!t]
     \centering
     \begin{subfigure}[t]{0.2\linewidth}
         \centering
         \includegraphics[width=\linewidth]{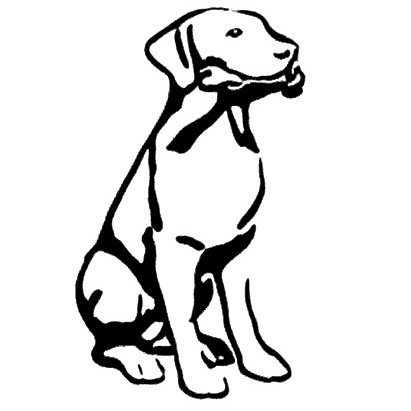}
     \end{subfigure}
     \begin{subfigure}[t]{0.2\linewidth}
         \centering
         \includegraphics[width=\linewidth]{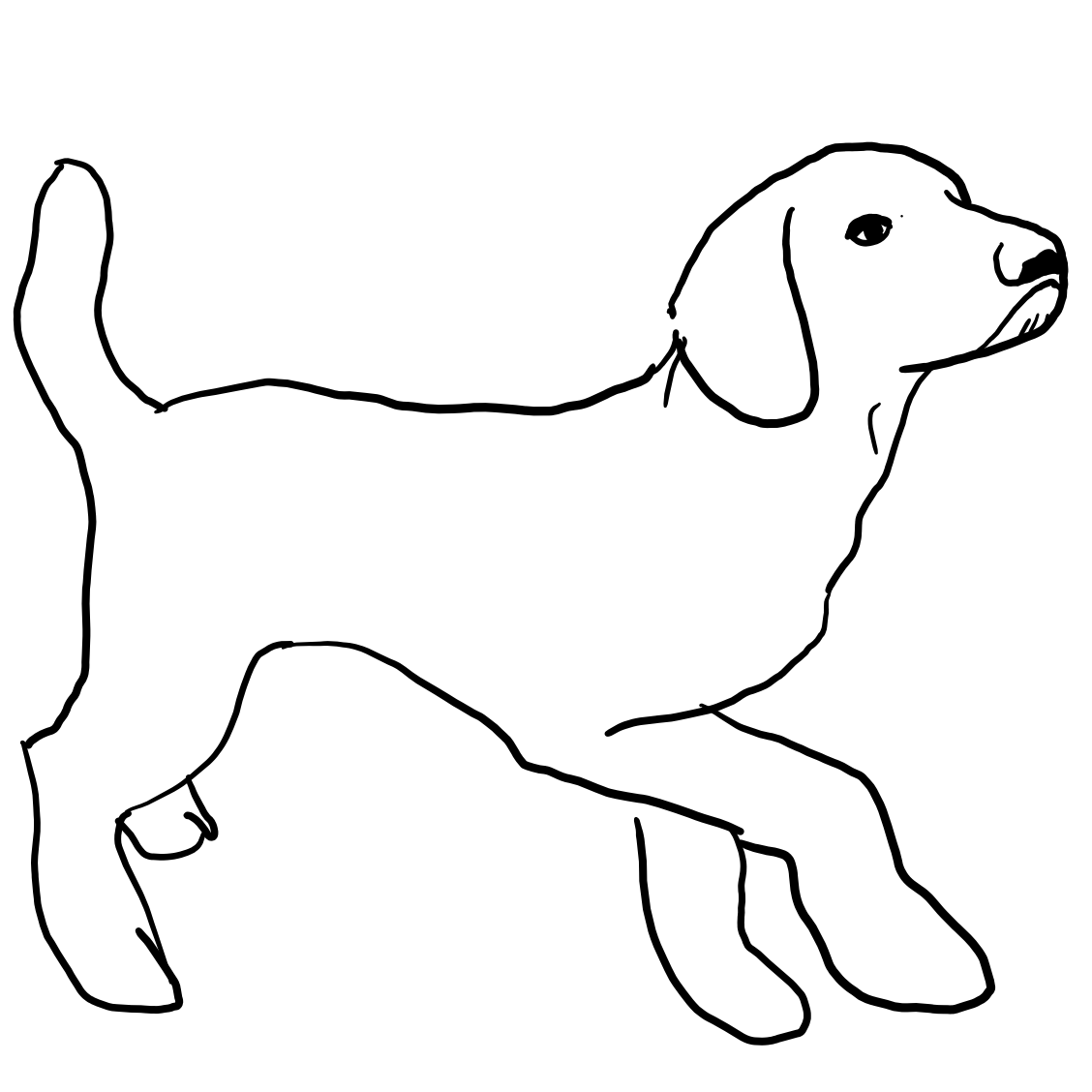}
     \end{subfigure}
     \begin{subfigure}[t]{0.2\linewidth}
         \centering
         \includegraphics[width=\linewidth]{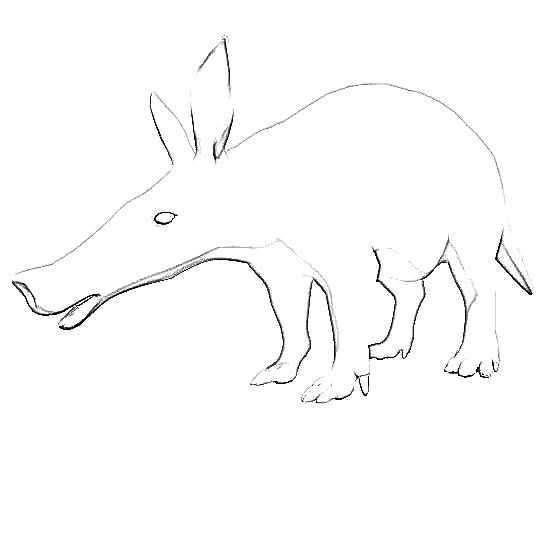}
     \end{subfigure}
     \begin{subfigure}[t]{0.2\linewidth}
         \centering
         \includegraphics[width=\linewidth]{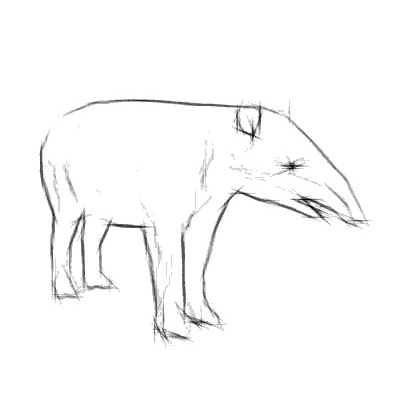}
     \end{subfigure}
     \begin{subfigure}[t]{0.2\linewidth}
         \centering
         \includegraphics[width=\linewidth]{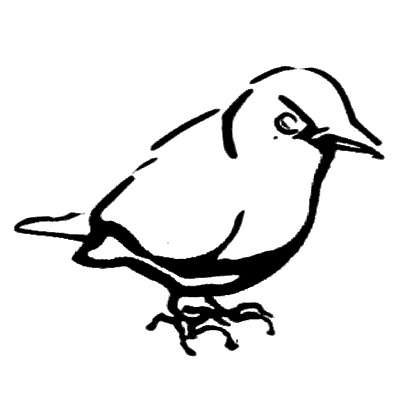}
     \end{subfigure}
     \begin{subfigure}[t]{0.2\linewidth}
         \centering
         \includegraphics[width=\linewidth]{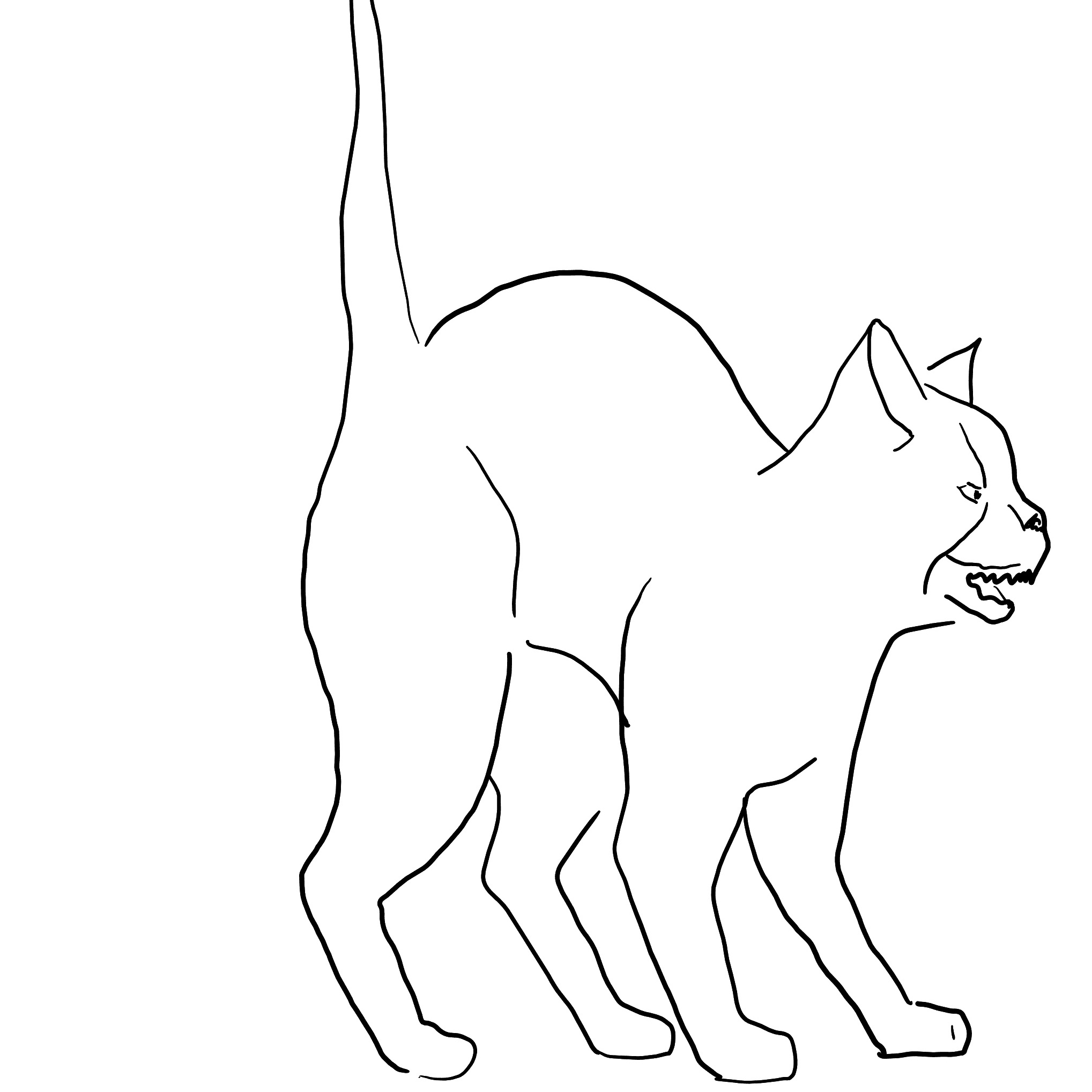}
     \end{subfigure}
     \begin{subfigure}[t]{0.2\linewidth}
         \centering
         \includegraphics[width=\linewidth]{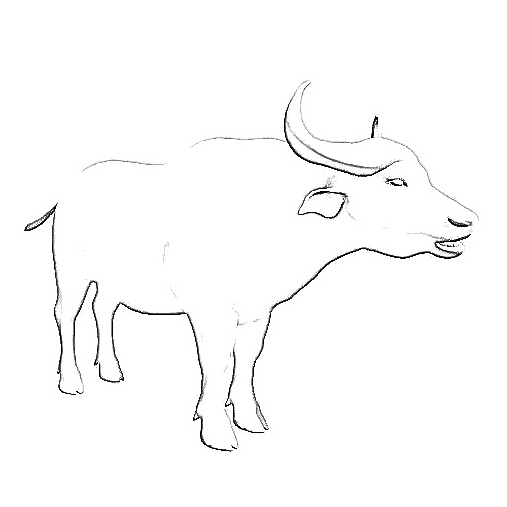}
     \end{subfigure}
     \begin{subfigure}[t]{0.2\linewidth}
         \centering
         \includegraphics[width=\linewidth]{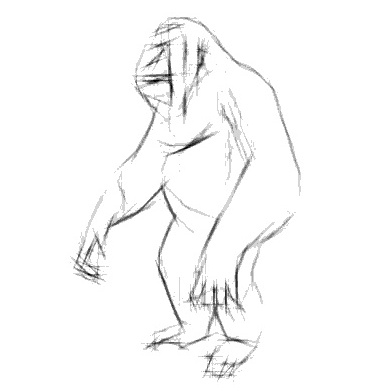}
     \end{subfigure}
    \caption{Sketch ground-truth in ANIMAR dataset, including various sketch types and animal poses.}
    \label{fig:sketch_gt}
\end{figure*}

Second, participants in our track challenge must solve the considerable domain gap between sketches and 3D shapes when dealing with differently posed animals. Furthermore, human sketches on existing datasets tend to be semi-photorealistic and drawn by experts. In contrast, our dataset comprises more diverse sketches, including abstract sketches drawn by amateurs, semi-photorealistic sketches, and sketches in different styles (\cf~Fig.~\ref{fig:sketch_gt}). As such, this task proves significantly more challenging than conventional sketch-based object retrieval tasks. We anticipate that the sketch-based 3D animal fine-grained retrieval task can pave the way for a new research direction and exciting, practical applications.

The remainder of this paper is organized as follows. Section~\ref{sec:related_work} provides an overview of related work. Section~\ref{sec:dataset} presents the ANIMAR dataset and the evaluation measures used in this SHREC contest. Section~\ref{sec:participants} describes the participant statistics. In Section~\ref{sec:methods}, the methods of the participating teams are presented. The evaluation results and an in-depth analysis of their performance are reported in Section~\ref{sec:results}. Finally, Section~\ref{sec:conclusion} concludes the paper and suggests directions for future work.

\begin{table}[!t]
\centering
\caption{SHREC challenge tracks for 3D object retrieval.}
\label{tab:3d_object_retrieval_tasks}
\resizebox{\columnwidth}{!}{%
\begin{tabular}{lcccc}
\hline
  \begin{tabular}[c]{@{}c@{}}\textbf{SHREC}\\\textbf{Challenge}\end{tabular} &
  \textbf{Year} &
  \begin{tabular}[c]{@{}c@{}}\textbf{Query}\\\textbf{Type}\end{tabular} &
  \begin{tabular}[c]{@{}c@{}}\textbf{Training}\\\textbf{Category}\end{tabular} &
  \begin{tabular}[c]{@{}c@{}}\textbf{Testing}\\\textbf{Category}\end{tabular}  \\ \hline

Hameed~\etal~\cite{Hameed-SHREC2018} & 2018 & Image & Seen & Seen \\ 
Hameed~\etal~\cite{Hameed-SHREC2019} & 2019 & Image & Seen & Seen \\ 
Li~\etal~\cite{Li-SHREC2019} & 2019 & Image & Seen & Seen \\ 
Li~\etal~\cite{Li-SHREC2020} & 2020 & Image & Seen & Seen \\ 
Feng~\etal~\cite{Feng-SHREC2022} & 2022 & Image & Seen & Unseen \\ 
\highlight{Li~\etal~\cite{Li-SHREC2012}} & 2012 & Sketch & Seen & Seen \\ 
\highlight{Li~\etal~\cite{Li-SHREC2013}} & 2013 & Sketch & Seen & Seen \\ 
\highlight{Li~\etal~\cite{Li-SHREC2014}} & 2014 & Sketch & Seen & Seen \\ 
Juefei~\etal~\cite{Juefei-SHREC2018} & 2018 & Sketch & Seen & Seen \\ 
Juefei~\etal~\cite{Juefei-SHREC2019} & 2019 & Sketch & Seen & Seen \\ 
Qin~\etal~\cite{Qin-SHREC2022} & 2022 & Sketch & Seen & Seen \\ 

\rowcolor{lightgray} SketchANIMAR & 2023 & Sketch & Unseen & Unseen \\ \hline
\end{tabular}%
}
\end{table}

\section{Related Work}
\label{sec:related_work}

To recover 3D objects from a database, content-based 3D object retrieval examines the visual contents of the objects, such as color, texture, form, and geometric aspects. Many tracks concentrating on similar problems have been held in previous SHREC competitions (see Table~\ref{tab:3d_object_retrieval_tasks}) to promote research on content-based 3D object retrieval.

Several SHREC tracks concentrate on retrieving 3D items in a database that resemble the 3D objects used as a query. The attractiveness of sketch-based 3D object retrieval, in particular, stems from the organic and intuitive quality of freehand sketches, and it has garnered much attention in recent years. 

\highlight{Li~\etal~\cite{Li-SHREC2012, Li-SHREC2013, Li-SHREC2014} promoted this intriguing study by organizing SHREC tracks of sketch-based 3D shape retrieval. At that time, deep learning was not popular; thus, submitted solutions were based on hand-crafted features such as Scale-Invariant Feature Transform (SIFT), Histogram of Oriented Gradient (HOG), fourier descriptors, bag-of-features, and sparse coding. After that, Juefei~\etal~\cite{Juefei-SHREC2018, Juefei-SHREC2019} extended the task to 2D scene sketch-based 3D scene retrieval.} Domain adaptation algorithms, such as two-stream CNN with triplet loss, adversarial training, and different data augmentation techniques, were used to resolve the disagreement between two domains (\ie, sketch and 3D object). In addition, a competition for sketch-based 3D form retrieval in the wild was conducted by Qin~\etal~\cite{Qin-SHREC2022}, further advancing the task. They used a variety of 3D forms, including models created by scanning genuine objects, as well as large-scale sketches created by amateur artists with a range of sketching abilities. Furthermore, technologies, such as point cloud and multi-view learning using various deep learning architectures, were created to emulate actual retrieval circumstances.

Sketch-based 3D object retrieval methods can be grouped into two categories: model-based and view-based approaches. Model-based methods commonly utilize 3D CNN to extract 3D shape features directly from the original 3D representations. In the view-based approach, 2D convolutional neural networks (CNN) are frequently used to analyze shape features from a set of 2D view projections.

Regarding the model-based approach, Furuya \etal~\cite{Furuya-BMVC2016} propose Deep Local Feature Aggregation Network (DLAN), which extracts rotation-invariant 3D local features and aggregates them in a single deep architecture. More concretely, the DLAN uses a set of 3D geometric features invariant to local rotation to characterize local 3D regions of a 3D model. The DLAN then compiles the set of features into a (global) rotation-invariant and compact feature for each 3D model. Furthermore, an Octree-based Convolutional Neural Network (O-CNN)~\cite{Wang-TransGraph2017} is also proposed for 3D shape analysis. O-CNN executes 3D CNN operations on the octants filled by the 3D shape surface using the average normal vectors of a 3D model sampled in the smallest leaf octants as input.

Concerning the view-based approach, Wang \etal~\cite{Wang-CVPR2015} propose two Siamese Convolutional Neural Networks for the views and the sketches. Moreover, the loss function is designed for within-domain and cross-domain similarities. Similarly, two deep CNNs are proposed by Xie \etal~\cite{Xie-CVPR2017} for deep feature extraction of sketches and 2D projections of 3D shapes. Next, the authors compute the Wasserstein barycenters of deep features of multiple projections of 3D shapes to form a barycentric representation. Last but not least, Multi-view Convolutional Neural Network (MVCNN)~\cite{Su-ICCV2015} creates a single, compact shape descriptor from data from multiple views of a 3D shape, which improves recognition performance.

In recent competitions of 3D Shape Retrieval Contests (SHREC)~\cite{Savva-3DOR2017, Moscoco-SHREC2020, Qin-SHREC2022}, teams achieving high performances followed the view-based approach.

\section{Dataset and Evaluation}
\label{sec:dataset}

\subsection{Dataset}

In this competition, we constructed a new dataset, namely ANIMAR, which encompasses a corpus of 711 distinct 3D animal models along with 140 sketch queries.

We collected an assemblage of 186 mesh models depicting over 50 diverse categories of animals. These models were diligently sourced from an array of publicly available online resources and video games, including the well-known Planet Zoo video game\footnote{\url{https://www.planetzoogame.com}}~\cite{Wu-NeurIPS2022}. The primary goal of our competition track was to simulate real-life scenarios in which users endeavor to identify and explore a diverse range of animal species. To achieve this, \textbf{we purposely concealed categorical information during both the training and retrieval stages}. Furthermore, we refined our model database by generating a series of watertight mesh models by reducing the number of faces \highlightx{by $25\%, 50\%,$ and $75\%$}, yielding a total of 525 models. Following the work of Douze \etal~\cite{Douze-2021}, our 3D animal model database is employed for both the training and retrieval phases.

\highlight{From 186 original mesh models, we randomly selected 60 models for sketch image creation. For each model, we rotated the model and generated 2-3 sketches from distinct viewpoints, thereby producing a total of 140 sketch images to describe the 3D animal models. Notably, we intentionally chose not to create sketches for all animal models in order to prevent participants from utilizing them to train retrieval solutions. Of the 140 sketches, 74 were aligned with their corresponding models in the database, yielding a set of 297 query-model pairs that were utilized for training purposes. The remaining 66 sketches were designated as queries, resulting in 265 query-model pairs employed during the retrieval phase. Unlike existing datasets, which primarily featured semi-photorealistic sketches drawn by experts, \textbf{our dataset comprises more diverse sketches, including abstract sketches drawn by amateurs, semi-photorealistic sketches, and sketches in different styles.} This diversity is exemplified in Fig.~\ref{fig:sketch_gt}, where the varied nature of the sketches can be observed.}

\subsection{Evaluation Metrics}

\highlightx{We provide a comprehensive evaluation of the performance of different methods in this track. The following metrics are utilized:}
\begin{itemize}
    \item \textbf{Nearest Neighbor (NN)} evaluates top-1 retrieval accuracy.
    \item \textbf{Precision-at-10 (P@10)} is the ratio of relevant items in the top-10 returned results.   
    \item \textbf{Normalized Discounted Cumulative Gain (NDCG)} is a measure of ranking quality defined as $\sum_{i=1}^{p} \frac{rel_i}{log_2(i + 1)}$, where $p$ is the length of the returned rank list, and $rel_i$ denotes the relevance of the i-th item.  
    \item \textbf{Mean Average Precision (mAP)} is the area under the precision-recall \highlightx{curve. It measures} the precision of methods at different levels and then takes the average. mAP is calculated as $ \frac{1}{r} \sum_{i=1}^{r} P(i)(R(i) - R(i-1))$, where $r$ is the \highlightx{number} of retrieved relevant items, $P(i)$ and $R(i)$ are the precision and recall at the position of the $i^{th}$ relevant item, respectively.
    \item \highlight{\textbf{First Tier (FT)} denotes the recall of the top $m$ retrieval results, where $m$ is the number of relevant images in the whole database. It measures the accuracy of retrieving the most relevant images among all the possible matches. The FT score is calculated as: FT = (number of relevant images retrieved in the top $m$) / $m$.}
    \item \highlight{\textbf{Second Tier (ST)} denotes the recall of the top $2m$ retrieval results, where $m$ is the number of relevant images in the whole database. It measures the ability to retrieve relevant images within a broader set of results. The ST score is calculated as: ST = (number of relevant images retrieved in the top $2m$) / $m$.}
    \item \highlight{\textbf{Fallout Rate (FR)} shows the ratio of non-relevant retrieved items in relation to the total number of non-relevant items available. It measures the system's ability to avoid retrieving non-relevant items. The FR score is calculated using the formula: FR = (number of non-relevant items retrieved) / (number of total non-relevant items in the database)}
\end{itemize}

\section{Participants}
\label{sec:participants}

Eight groups participated in the SketchANIMAR challenge track. Each group was provided with three weeks to complete the challenge. Throughout the contest, a total of 204 runs were submitted. All participating groups were required to register and submit their results along with a detailed description of their methods. It is important to note that the organizers did not participate in the challenge. \highlight{We remark that three teams opted not to disclose the methods they used in the competition against the SHREC spirit, which was born to compare the performance of algorithms on common data. Thus, they \highlightx{are not} reported in this paper.} The participant details are provided below \highlight{(team members will be added upon acceptance)}:

\begin{itemize}
    \item TikTorch team submitted by Nhat-Quynh Le-Pham, Huu-Phuc Pham, Trong-Vu Hoang, Quang-Binh Nguyen, and Hai-Dang Nguyen 
    (see Section \ref{sec:team_tiktorch}). 
    \item THP team submitted by Truong Hoai Phong 
    (see Section \ref{sec:team_thp}).
    \item Etinifni team submitted by Tuong-Nghiem Diep, Khanh-Duy Ho, Xuan-Hieu Nguyen, Thien-Phuc Tran, Tuan-Anh Yang, Kim-Phat Tran, Nhu-Vinh Hoang, and Minh-Quang Nguyen 
    (see Section \ref{sec:team_etinifni}). 
    \item V1olet team submitted by Trong-Hieu Nguyen-Mau, Tuan-Luc Huynh, Thanh-Danh Le, Ngoc-Linh Nguyen-Ha, and Tuong-Vy Truong-Thuy 
    (see Section \ref{sec:team_v1olet}).
    \item DH team submitted by Hoai-Danh Vo and Minh-Hoa Doan 
    (see Section \ref{sec:team_dh}).
\end{itemize}

\section{Methods}
\label{sec:methods}


\subsection{Overview of Submitted Solutions}

\begin{figure}[!t]
	\centering
 \includegraphics[width=\linewidth]{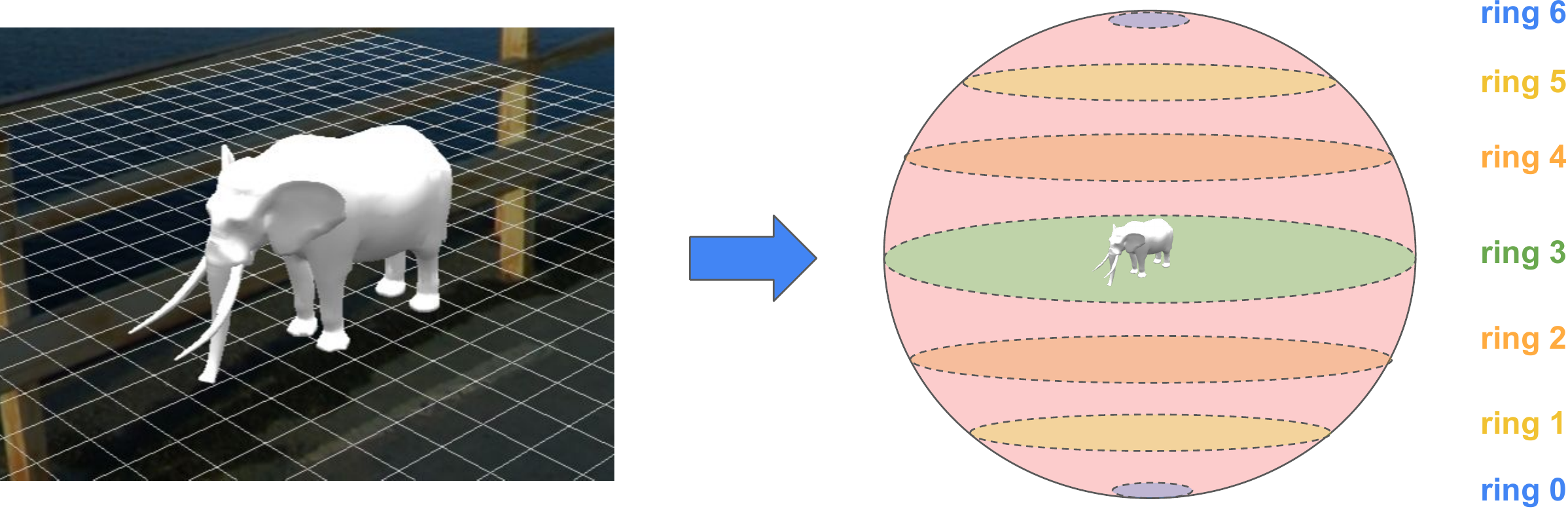}
        \caption{3D object represented as view sequences of 7 rings with 12 views on each ring. The chosen latitudes were $0$ (the equator), $\pm 90$ (the poles), and $\pm 30, \pm 60$.}
	\label{fig:multiviewrepresentation}
\end{figure}

\highlightx{All submissions to} our track are built upon the foundation of view-based learning. This approach captures the essence of each 3D object by presenting it as a sequence of ring images, as illustrated in Fig.~\ref{fig:multiviewrepresentation}. These images are acquired by strategically maneuvering a camera around the object along a predefined path, with each ring consisting of a series of images. In particular, when the camera's trajectory aligns parallel to the ground plane relative to the object, the multi-view method demonstrates remarkable effectiveness, generating valuable images that greatly assist in extracting features for representing three-dimensional objects.

\begin{figure*}[!t]
     \centering
     \includegraphics[width=\linewidth]{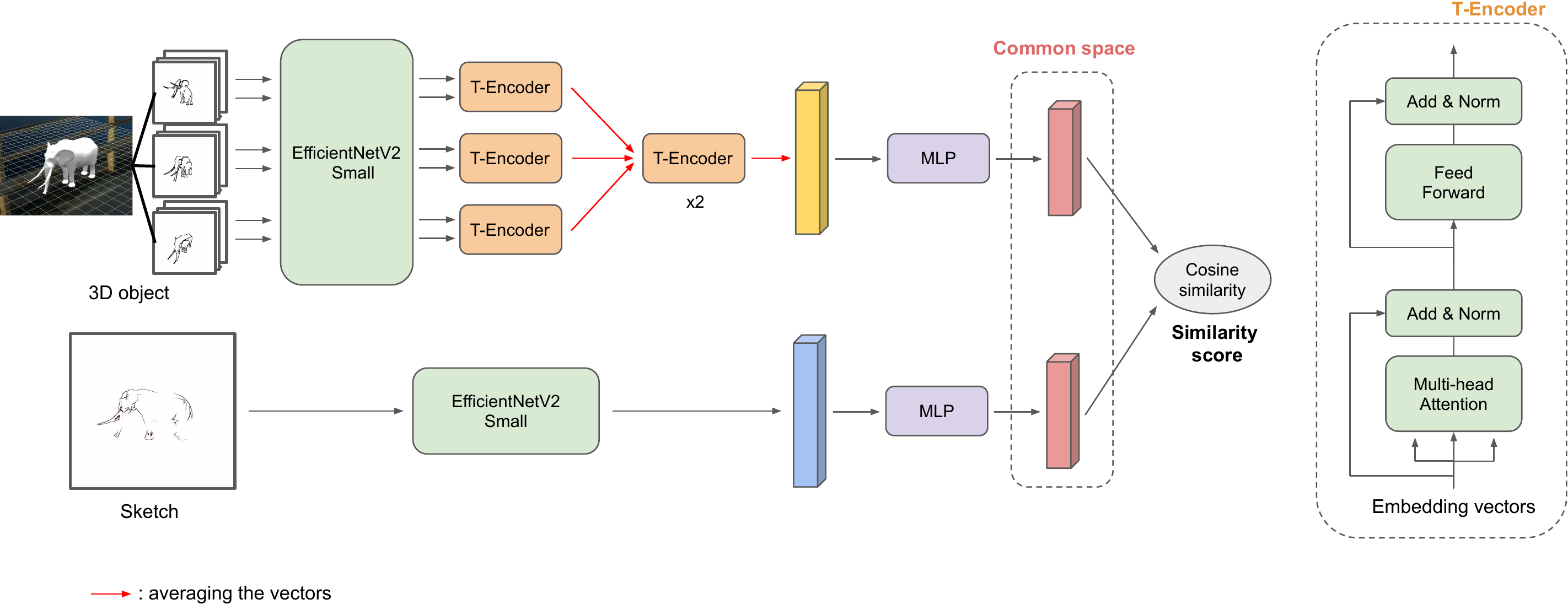}
     \caption{Proposed framework of TikTorch team.}
     \label{fig:overview_tikTorch}
\end{figure*}

The view-based learning shows more advantages than directly learning the point clouds. This matches well with the settings of our challenge. \highlight{The 3D objects in our dataset have high-density point clouds with a large number of points.} This detail can make it difficult for feature extraction models on point cloud such as PointNet \cite{Qi-CVPR2017} and PointMLP \cite{Ma-2022} when these models usually randomly sample a specific number of points (1024, for example) in the point cloud of 3D objects. This approach proves particularly useful in scenarios where 3D models \highlight{are not readily available} for querying, but sketches of the objects are, which is often the case in real-world applications.

To facilitate the retrieval task, TikTorch, THP, and Etinifni teams considered the problem as contrastive learning (as shown in Sections~\ref{sec:team_tiktorch},~\ref{sec:team_thp},~\ref{sec:team_etinifni}). Meanwhile, V1olet team formulated the task as a classical classification problem (as depicted in Section~\ref{sec:team_v1olet}). On the other hand, DH team directly extracted and compared non-deep learning features between sketch queries and generated sketches from 3D objects (see Section~\ref{sec:team_dh}).


\subsection{TikTorch Team}
\label{sec:team_tiktorch}

\subsubsection{Proposed Contrastive Learning Solution}

To retrieve 3D objects from sketch queries, they propose a contrastive learning framework where embedding vectors of 3D objects and 2D sketches are learned. The embedding vectors of similar objects and sketches should be closer to each other and vice versa. 

The overall architecture of their method is presented in Fig.~\ref{fig:overview_tikTorch}, containing two separate feature extractors for 3D objects and sketch images. The extracted feature vectors are then embedded in the common vector space by two Multi-layer Perceptron (MLP) networks. The contrastive loss used for simultaneous learning of the parameters for models is a customized version of Normalized Temperature-scaled Cross Entropy Loss (NT-Xent)~\cite{Chen-ICML2020}.

\textbf{Sketch feature extractor.} To extract the features of sketch images, they fine-tune EfficientNetV2-Small~\cite{Tan-ICML2021} pretrained on ImageNet dataset~\cite{Russakovsky-IJCV2015}. The models in the EfficientNetV2 family reduce the parameter size significantly while maintaining competitive accuracy on many datasets, which is desirable for simple images, especially sketch images.

\textbf{3D object feature extractor.} Each 3D object is represented as a set of 3 rings, and each ring contains 12 images. The 3D object feature extractor has two main phases: extracting the features of each ring (ring extractor) and combining the features of 3 rings to obtain the features of the object.

In the ring extractor, they also fine-tune EfficientNetV2-Small~\cite{Tan-ICML2021}, similar to the sketch feature extractor module, to extract the features of 12 images of each ring. These 12 feature vectors then go through an encoder block called T-Encoder~\cite{Vaswani-NeurIPS2017} to learn the relationship between images in the same ring to decide which image is essential in the current ring and which is not. After that, they combine these vectors by simply calculating their average to get a single feature vector for each ring.

When they obtain the feature vectors of 3 rings, these vectors are passed into 2 T-encoder blocks to know which ring is useful for the model to learn the features of the current object. Then, the vectors are averaged to get the feature vector of the 3D object.

\textbf{Embedding into common space.} To compute the similarity between objects and sketches, their feature vectors must be embedded into a shared space. Since the feature vectors of 3D objects and sketches may have different dimensions, two MLP networks with two layers are utilized. The output layer of each network has the same number of units, ensuring that the feature vectors are transformed into the same vector space. In addition, a Dropout layer~\cite{Hinton-2012} is added to each network to prevent overfitting. Once the feature vectors are embedded in the common space, the similarity between two embedding vectors, $\mathbf{u}$ and $\mathbf{v}$, can be computed using the cosine similarity metric.

\begin{equation}
    \text{sim}\left (\mathbf{u}, \mathbf{v} \right ) = \frac{\mathbf{u} \cdot \mathbf{v}}{\left \| \mathbf{u} \right \| \left \| \mathbf{v} \right \|}
    \label{eq:cosine}
\end{equation}

\textbf{Loss function.} The contrastive loss function used is a customized version of Normalized Temperature-scaled Cross Entropy Loss (NT-Xent)~\cite{Chen-ICML2020}. Given a mini batch of $2N$ samples $\{\mathbf{x}_i\}_{i=1}^{2N}$ containing $N$ objects and $N$ sketches. They denote $\mathbf{z}_i$ as the embedding vector of the sample $\mathbf{x}_i$ in the common space. Let $P_i$ be the set of indices of samples that are similar to $\mathbf{x}_i$ in the current mini-batch exclusive of $i$, \ie, $(\mathbf{x}_i, \mathbf{x}_j)$ is a positive pair for $j \in P_i$. Here, $\mathbf{x}_i$ can belong to many positive pairs, such as two 3D objects that are similar to the same sketch. The loss function for a positive pair $(\mathbf{x}_i, \mathbf{x}_j)$ is defined as:

\begin{equation}
    l_{i,j} = -\log\frac{\exp\left(\text{sim}\left(\mathbf{z}_{i}, \mathbf{z}_{j}\right)/\tau\right)}{\sum^{2N}_{k=1} \mathbb{I}_{[k\neq{i}, k \notin P_i]}\exp\left(\text{sim}\left(\mathbf{z}_{i}, \mathbf{z}_{k}\right)/\tau\right)},
\end{equation}
where $\mathbb{I}_{[k\neq{i}, k \notin P_i]} \in \{ 0,1 \}$ is an indicator function evaluating to 1 if and only if $k \neq i$ and $k \notin P_i$, $\tau$ is a temperature parameter.

\textbf{Training phase.} During the training process, the optimizer used for training was AdamW~\cite{Loshchilov-2017}, along with the StepLR algorithm to reduce the learning. They also applied the $k$-fold cross-validation technique with $k = 5$.

\textbf{Retrieval phase.} They ensemble the results of models trained on $k$-fold by max-voting. The similarity between a 3D object and a sketch image is the largest value of the similarity score computed by the five models.

\subsubsection{Data Augmentation}

\begin{figure}[!t]
	\centering
	\includegraphics[width=\linewidth]{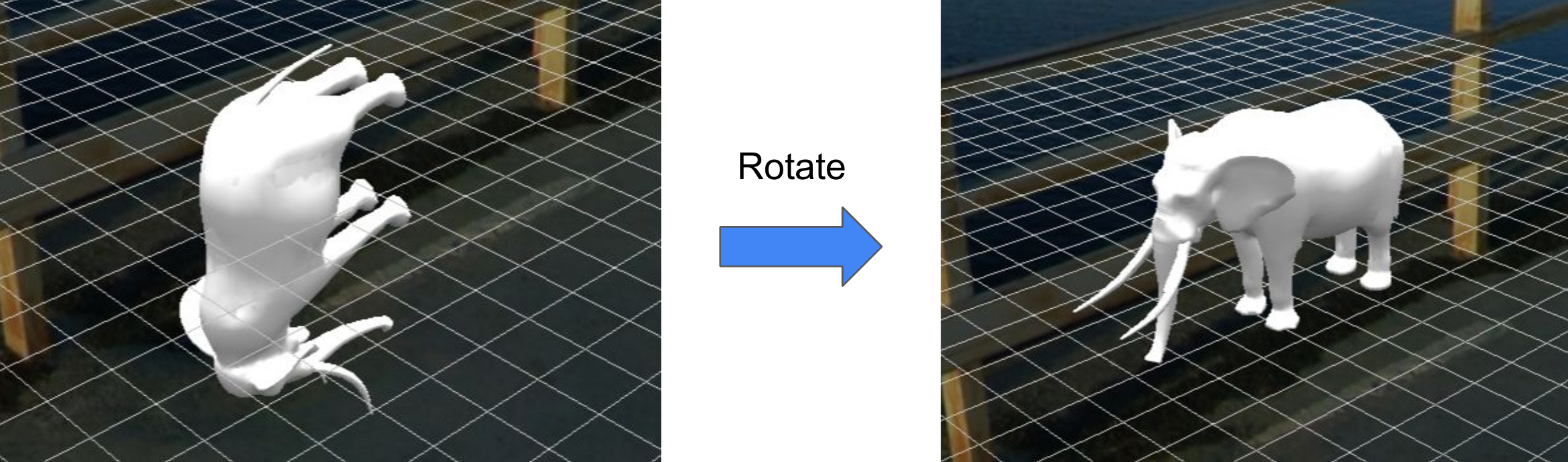}
	\caption{Example of rotating a 3D object whose axis is not aligned with the majority of objects.}
	\label{fig:objectrotation}
\end{figure}

\textbf{Generation of multi-view images for 3D objects.} Before generating batch images from 3D objects, it is essential to ensure axial synchronization of the objects so that the resulting multi-view images with their corresponding camera angles are consistent. To achieve this, they carefully examine the available dataset and identify several objects rotated at a 90-degree angle along the Ox axis. Then, they apply a consistent rotation to align these objects with the majority of the dataset as in Fig.~\ref{fig:objectrotation}. 

Among seven rings, as in Fig.~\ref{fig:multiviewrepresentation}, they find that the most informative views are captured from rings 2, 3, and 4, which provide a 360-degree perspective around the object. Hence, they focus on processing the images from these rings to extract the relevant features and information.

\begin{figure}[!t]
	\centering
	\includegraphics[width=\linewidth]{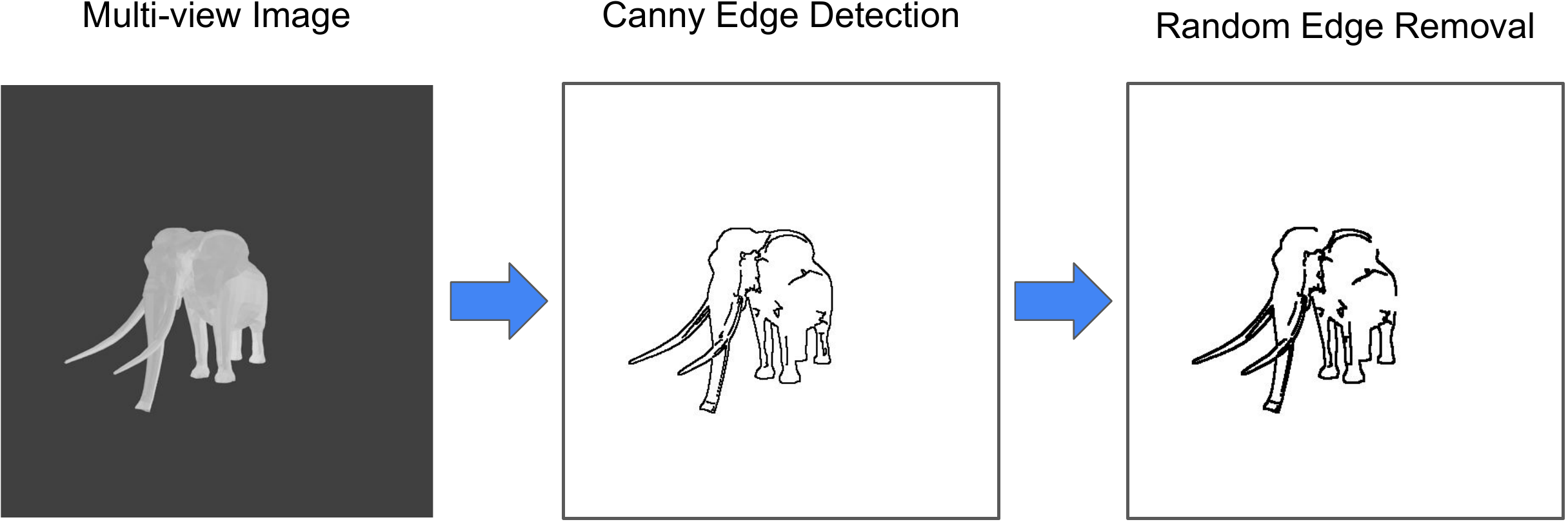}
	\caption{Multi-view image processing steps.}
	\label{fig:imageprocess}
\end{figure}

\begin{figure}[!t]
	\centering
	\includegraphics[width=\linewidth]{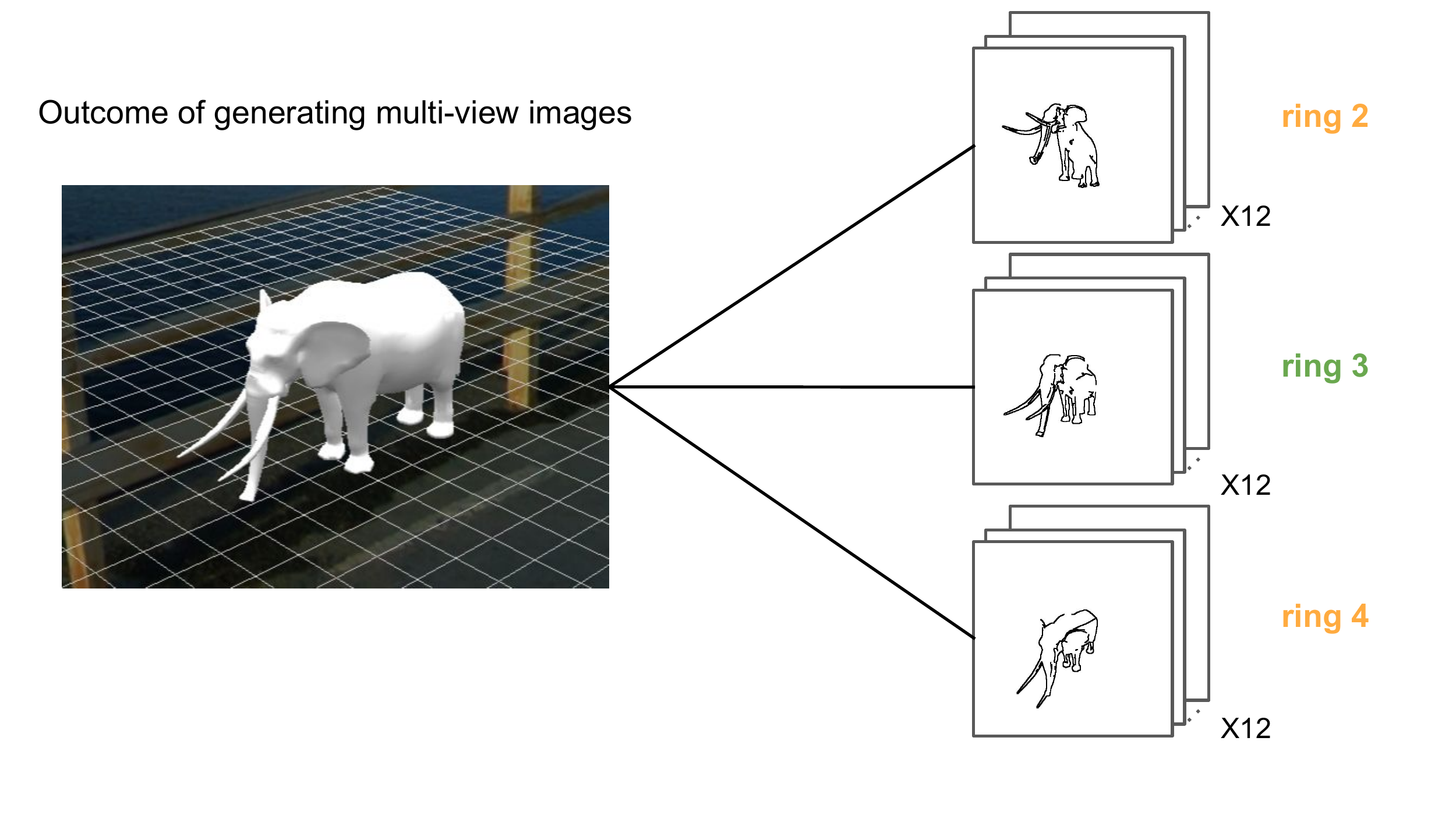}
	\caption{Outcome of generating multi-view images for 3D objects. Each 3D object is represented by a set of 3 rings, and each ring is a collection of 12 images.}
	\label{fig:finaloutcome}
\end{figure}

To enhance the sketch-like appearance of the multi-view images, they utilize the Canny edge detector~\cite{Canny-TPAMI1986} to extract edge information. They also add some noises and variations to the edge information by randomly removing edges in the image using a traversal algorithm while preserving the underlying structure and content of the image (see Fig.~\ref{fig:imageprocess}). Figure~\ref{fig:finaloutcome} illustrates the outcome of generating multi-view images for a 3D object.

\begin{figure}[!t]
	\centering
	\includegraphics[width=\linewidth]{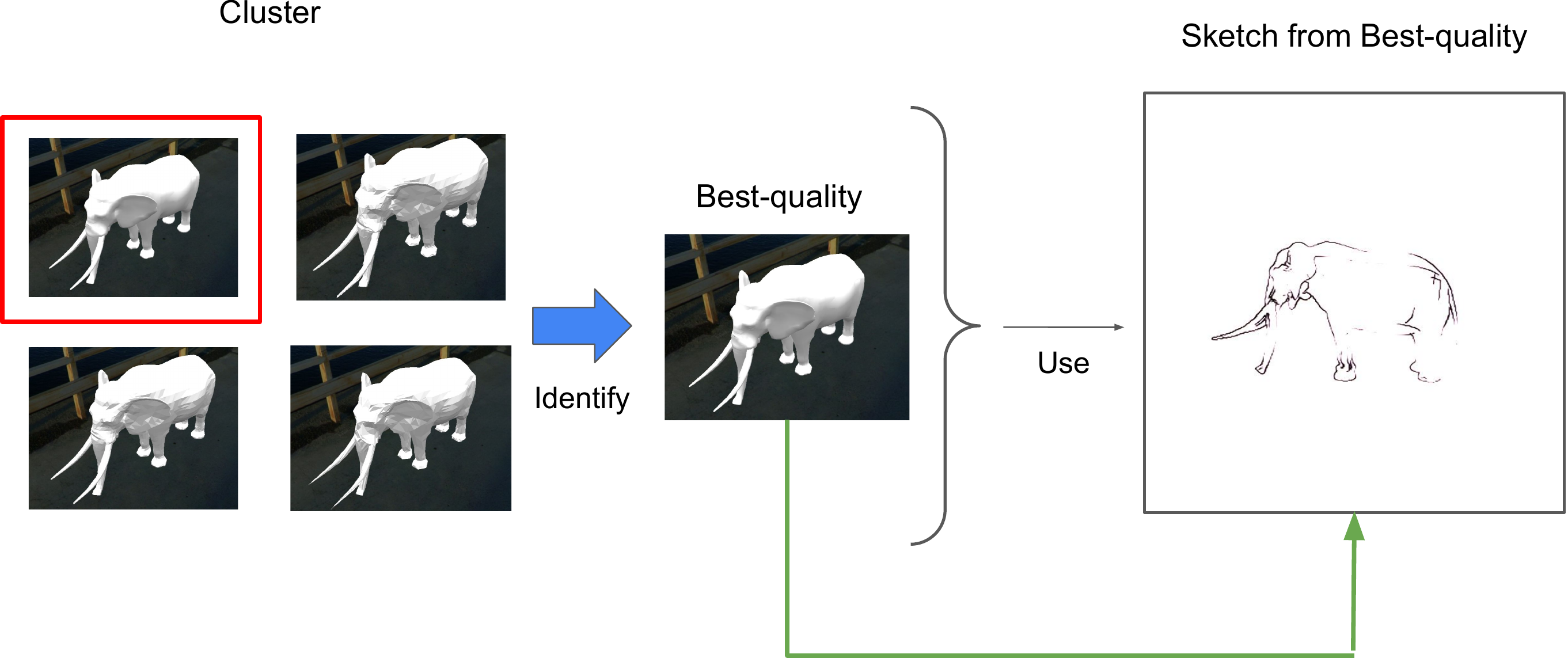}
	\caption{Optimizing object sketches through clustering: the sketch of the best-quality object for all objects within a group of similar 3D objects is used.} 
	\label{fig:clustersketch}
\end{figure}

\textbf{Generation of training sketch query images.} Firstly, they cluster similar 3D objects together by some algorithms to check the similarity between the distribution of points in the point clouds and also the manual checking as post-processing. After that, they identify the best-quality object in each cluster, which usually is the most fine-grained object (\ie, the highest number of points in the point cloud). Then, when they generate a sketch-line image for each object in this cluster and use it for contrastive learning, they pick the image of the best-quality object as the representative sketch (depicted in Fig.~\ref{fig:clustersketch}).

To expand training samples for contrastive learning, they develop a method to generate three queries per object. Each query is randomly chosen from rings 2, 3, and 4 (see Fig.~\ref{fig:multiviewrepresentation}) with probabilities of 0.2, 0.6, and 0.2, respectively, as they observe that the majority of informative queries are in ring 3. Once a ring is selected, they randomly choose an image within that ring from the cluster this object belongs to and apply random Canny edge~\cite{Canny-TPAMI1986} or Artline~\cite{Madhavan-2021} techniques, along with image horizontal flip and rotation transformations. By implementing this process, they can significantly increase the number of our training samples from about 100 to 2500 while maintaining a high level of quality.

\begin{figure*}[!t]
     \centering
     \includegraphics[width=0.85\linewidth]{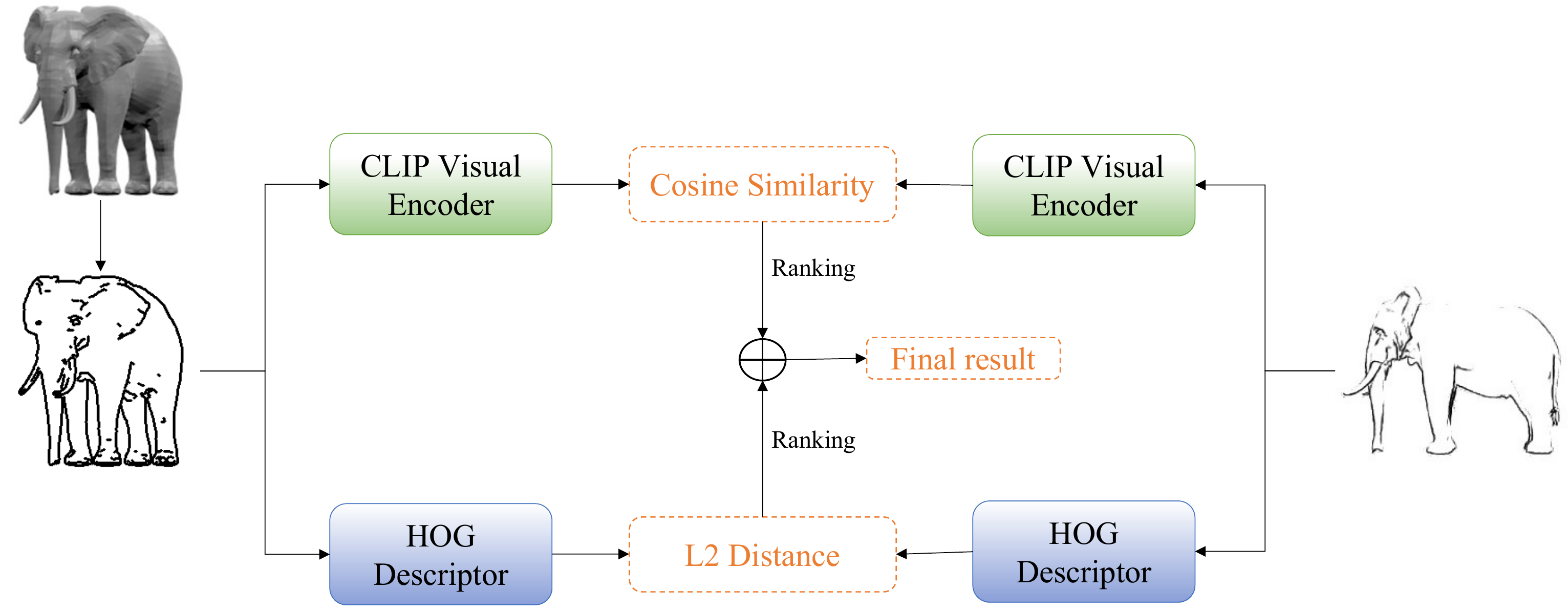}
     \caption{Proposed framework of THP team.}
     \label{fig:overview_thp}
\end{figure*}

\subsection{THP Team}
\label{sec:team_thp}

\subsubsection{Architecture of Proposed Network}

Fig.~\ref{fig:overview_thp} illustrates an overview of the proposed network. To evaluate the similarity of two given sketch images, they compare the distance of global features and local features and then combine the results. 

To extract global features, they use the pre-trained CLIP model~\cite{Radford-ICML2021}. The input of the CLIP model is the dilated Canny edge extracted from multi-view images. After that, CLIP feature vectors of view images and sketch queries are matched using cosine similarity. The final score is calculated as the maximum of scores of 4 views, in which each view score is calculated by the sum of the six highest similarities.

To increase local information awareness, they use the HOG descriptor~\cite{Dalal-CVPR2005} on both sketch and multi-view images. The HOG vectors are then matched using the L2 distance. They are also ranked like CLIP feature vectors.

Finally, they combine CLIP and HOG similarity scores as follows: \textit{Score = $\alpha$ * CLIP score + (1 - $\alpha$) * HOG score}, where $\alpha=0.7$.

\subsubsection{2D Shape Projection}

They use four camera setups to take multiple views of 3D objects:

\begin{itemize}
    \item For the first camera setup, assuming the 3D object is initially aligned along the $z$-axis, the camera is aligned on the $Oxy$ plane and looks at the center of the object. The camera is moved around the subject to create 12 views from a distance of 30 degrees each time.  

    \item For the second camera setup, the camera is raised to 30 degrees above the $Oxy$ plane and moved around to create the next 12 views. 

    \item For the third camera setup, the camera is placed on the Oyz plane and looks at the object's center. The camera is moved like the first setup to create the next 12 views. 

    \item Camera for the last setup is raised from the third setup to 30 degrees to create the next 12 views. 
\end{itemize}

There are a total of 48 views for each object. Note that it is possible to create images from other directions, but according to their tests, from these 48 views, they can observe the characteristics of objects.

\subsubsection{Sketch Pre-processing}

\begin{figure}[!t]
    \centering
    \includegraphics[width=\linewidth]{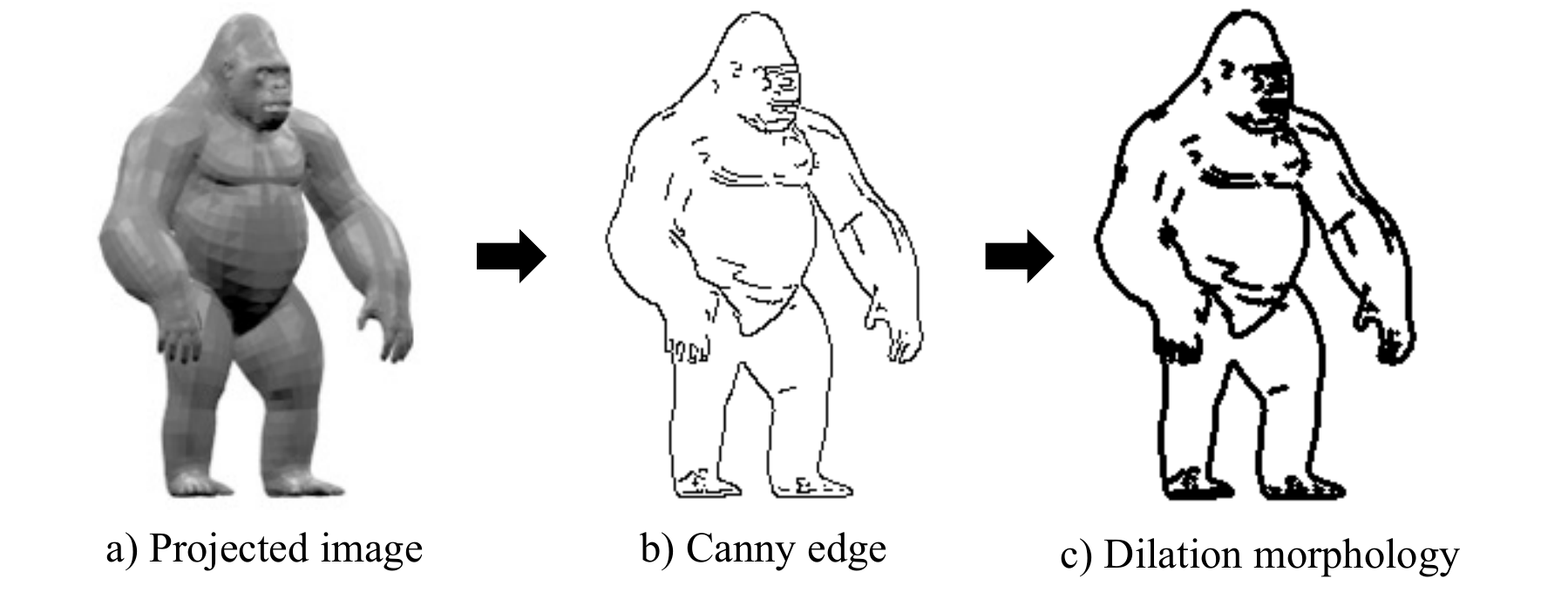}
    \caption{Pipeline of sketch pre-processing.}
    \label{fig:sketch_processing}
\end{figure}

To reduce the domain gap, they apply Canny edge algorithm~\cite{Canny-TPAMI1986} for each 2D projection image to create an image similar to the sketch image because the sketch is also a special type of edge. Both sketch and Canny edge images are then cropped and resized to $224 \times 224$ with the padding of 5 pixels. After that, edges are further clarified using the dilation morphology algorithm. The sketch pre-processing pipeline is shown in Fig.~\ref{fig:sketch_processing}.

\subsection{Etinifni Team}
\label{sec:team_etinifni}

\begin{figure*}[!t]
     \centering
     \includegraphics[width=\linewidth]{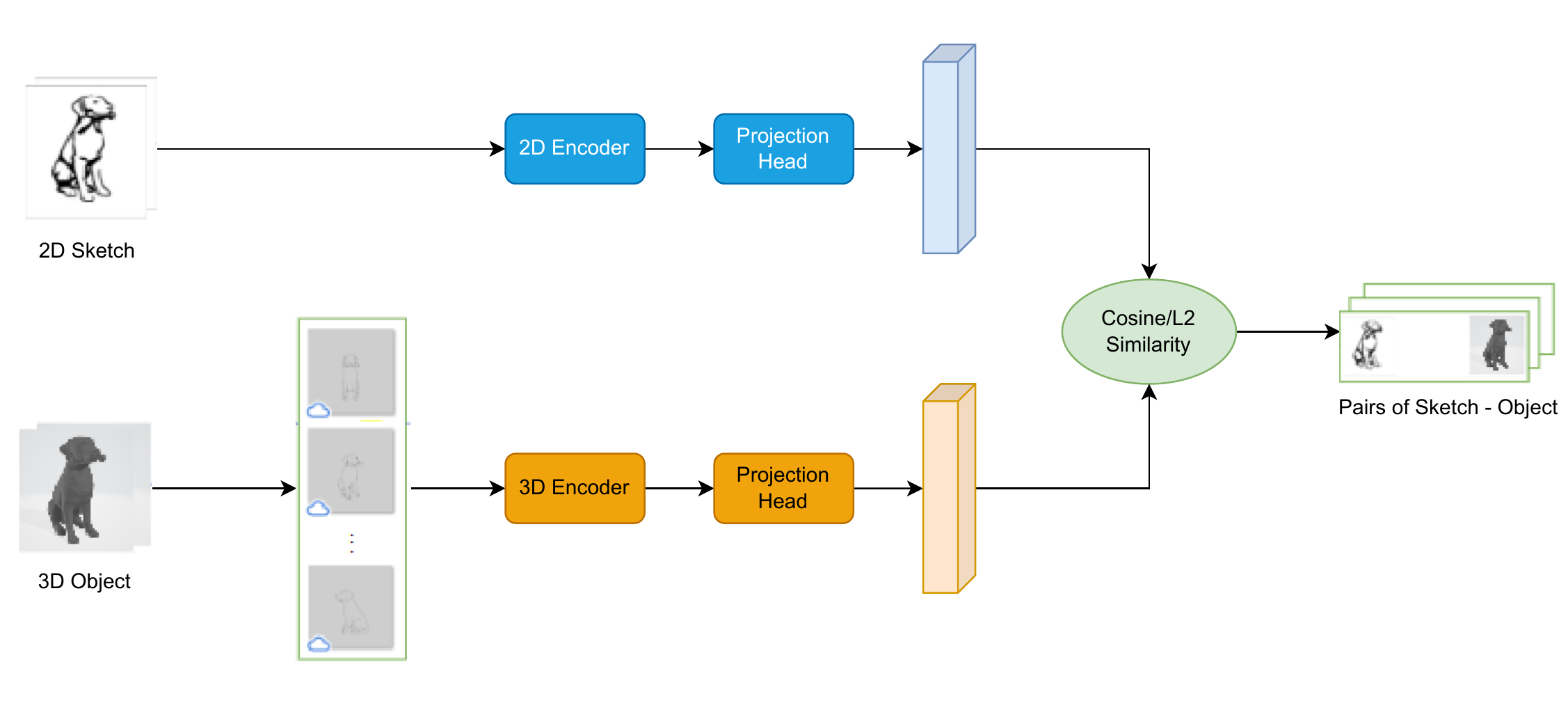}
     \caption{Proposed framework of Etinifni team.}
     \label{fig:overview_etinifni}
\end{figure*}

\subsubsection{Overview of Proposed Solution}

Due to the nature of the retrieval tasks, they build a deep learning framework using the CLIP model~\cite{Radford-ICML2021} as the backbone (See Fig.~\ref{fig:overview_etinifni}). The framework performs the following steps:
\begin{enumerate}
    \item Pre-process the dataset to re-direct the 3D objects into one single vertical orientation.
    \item Extract multi-views (\ie, 12 random views around the objects in uniform angles) from the 3D objects.
    \item Encode the 2D sketches and the view images of 3D objects using the AutoEncoder built upon the ResNet50.
    \item Reduce the size of the feature vectors from $512$ to $128$ through the projection head.
    \item Compare feature vectors of sketches and the 3D objects using the cosine similarity function. After that, they can identify the matching pairs of sketches and 3D objects.
\end{enumerate}

\subsubsection{Data Pre-Processing}

\textbf{Resize objects}: They resize the 3D objects to fit the 3D object inside an imaginary box of $2 \times 2 \times 2$ by re-scaling the dimensions $(x, y, z)$ of the 3D objects into the new dimension $(x', y', z')$ not greater than 2 using the following formula:
\begin{equation}
    (x', y', z') = \Biggl(\frac{2x}{\max{(x, y, z)}}, \frac{2y}{\max{(x, y, z)}}, \frac{2z}{\max{(x, y, z)}}\Biggl)\\
\end{equation}

\textbf{Re-orientate objects}: They re-direct the orientation of the 3D objects manually so that the 3D objects are standing (\ie,~the objects are in an erect position).

\subsubsection{Multi-view Generation}

They first rotate the camera as in Fig.~\ref{fig:multiviewrepresentation}. They apply 3D image rendering at each position to provide a more accurate and detailed view of objects. This enables the model to extract the necessary features for retrieval with greater precision, even where subtle variations in shape and texture can be crucial in determining an object's identity. Models then extract the most detailed and accurate information possible, leading to more robust and reliable results.

\subsubsection{Data Augmentation}

Due to the limited available training data, data augmentation techniques are applied to increase the number of training data:

\textbf{Outline modifications:}

\begin{itemize}
    \item \textit{Thickness:} is a significant attribute of 2D objects as it plays a crucial role in defining their physical properties and functionality. By adjusting the thickness, they generate new sketches with varying line lengths, thereby introducing diversity in the visual depiction of the objects.
    \item \textit{Geometry:} explores the properties and relationships of shapes, sizes, positions, and dimensions of objects in space. By modifying the stroke of these edges, they can generate variations in the appearance of the zigzag patterns. This allows for a deeper investigation into the geometric characteristics of the objects and the effects of stroke adjustments on their overall geometry.
\end{itemize}

\textbf{Image processing:}
\begin{itemize}
    \item \textit{Random deletion:} To ensure the robustness of the sketch, they partition the 2D images into multiple blocks. Each block is equal in size and contains a subset of pixels from the original image. They randomly select a portion of these blocks and remove them from the image to simulate missing strokes on the data set.
    \item \textit{Image compression:} Because the provided query images are of low quality, a compressor is utilized to reduce the image quality to generate the sketches.
\end{itemize}

\textbf{View extraction:}
\begin{itemize}
\item The camera is set up at a suitable distance, height, and orientation to ensure comprehensive coverage of the object's surface. The camera views are then randomly selected to capture diverse angles for robust 3D reconstruction.
\end{itemize}


\begin{figure*}[t!]
    \centering
    \includegraphics[width=\linewidth]{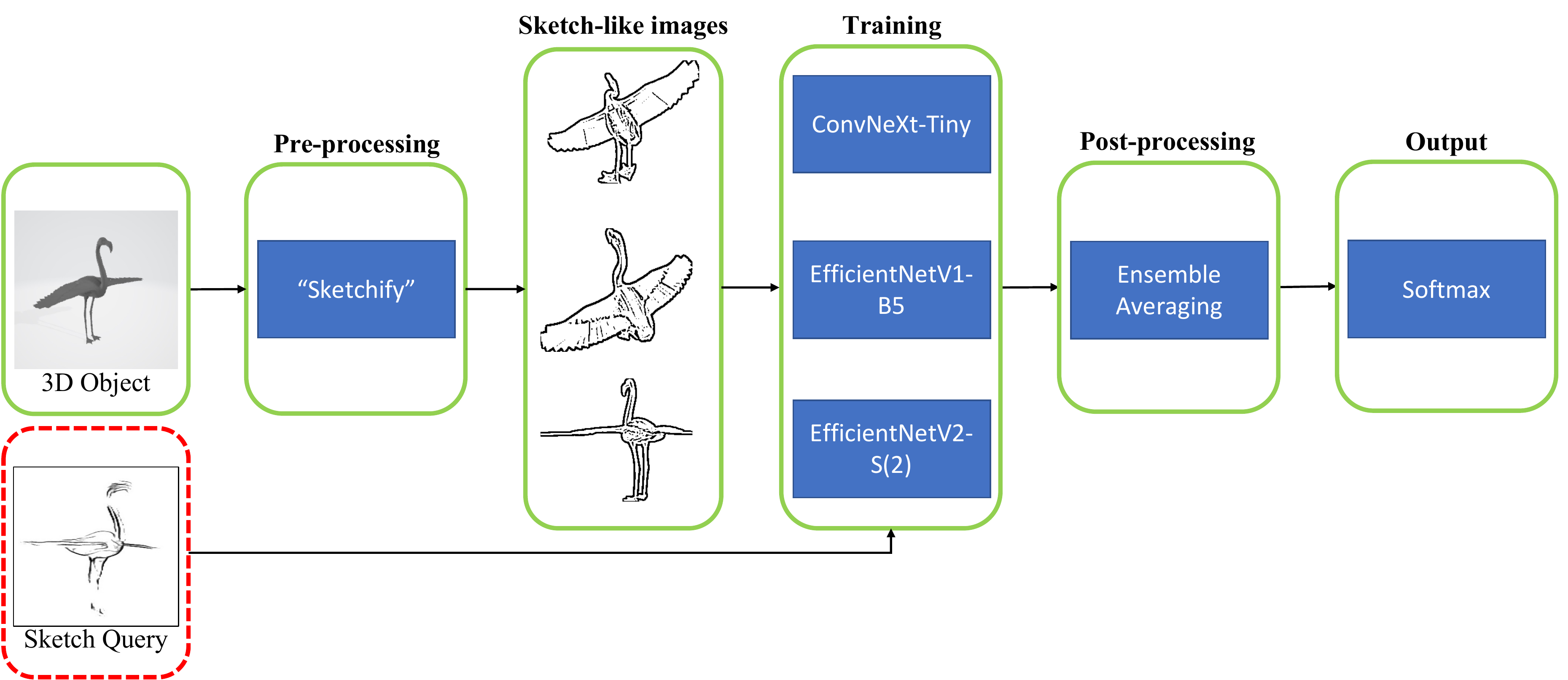}
    \caption{Proposed framework of V1olet team.}
    \label{fig:overview_v1olet}
    \vspace{-2mm}
\end{figure*}

\subsection{V1olet Team}
\label{sec:team_v1olet}

\subsubsection{Proposed Classification Approach}

They formulate the retrieval task as a classical classification problem. Notably, several potential CNNs, including EfficientNet \cite{Tan-ICML2019}, EfficientNetV2 \cite{Tan-ICML2021}, and ConvNeXt \cite{Liu-CVPR2022} are employed to recognize whether the sketch query and 3D objects are the same class. These networks are renowned for their remarkable feature extraction capabilities and are considered state-of-the-art in image recognition. The used models also are lightweight and suitable for real-life applications while achieving considerable performance.

\textbf{Ensemble Solution.} Figure~\ref{fig:overview_v1olet} illustrates the proposed ensemble approach by averaging the predictions of each model. This approach effortlessly helps mitigate individual models' potential shortcomings, resulting in improved performance and robust generalization to varying data distributions. 

To further improve the accuracy of models, they also utilize Test Time Augmentation (TTA), which involves applying a range of transformations such as rotations, flips, and translations to test images and averaging the results to obtain the final prediction. Specifically, they utilize horizontal flipping to provide additional perspectives of the original images. This technique not only enhances generalization but also enables the model to recognize objects that may be oriented differently from those in the training set.

\textbf{Training Phase.} To evaluate the performance of used models, they created a validation set by randomly leaving out 10\% of samples from each class in the training data. Pre-trained models on ImageNet were fine-tuned using the remaining training sets. They also employed the cross-entropy loss with label smoothing of 0.1 to prevent overfitting and improve the generalizability of models. Sketchifized multi-view images were jointly trained with the original sketch queries to enforce networks to recognize them as the same class. During both training and inference, an image size of $384\times 384$ was utilized. All networks were trained for 20 epochs using the Adam optimizer~\cite{Kingma-2014} with a learning rate of $0.0001$. Finally, they selected the models with the best validation accuracy for ensembling.

\textbf{Retrieval Phase.} Given a sketch query, the ensemble averaging of CNN models produces a set of softmax probabilities. These probabilities are then used to identify whether the sketch query and sketch-like images generated from the 3D object belong to the same class. The softmax probabilities also serve as a ranking metric, allowing for sorting the retrieved 3D objects by relevance.

\begin{figure}[t!]
    \centering
    \includegraphics[width=\linewidth]{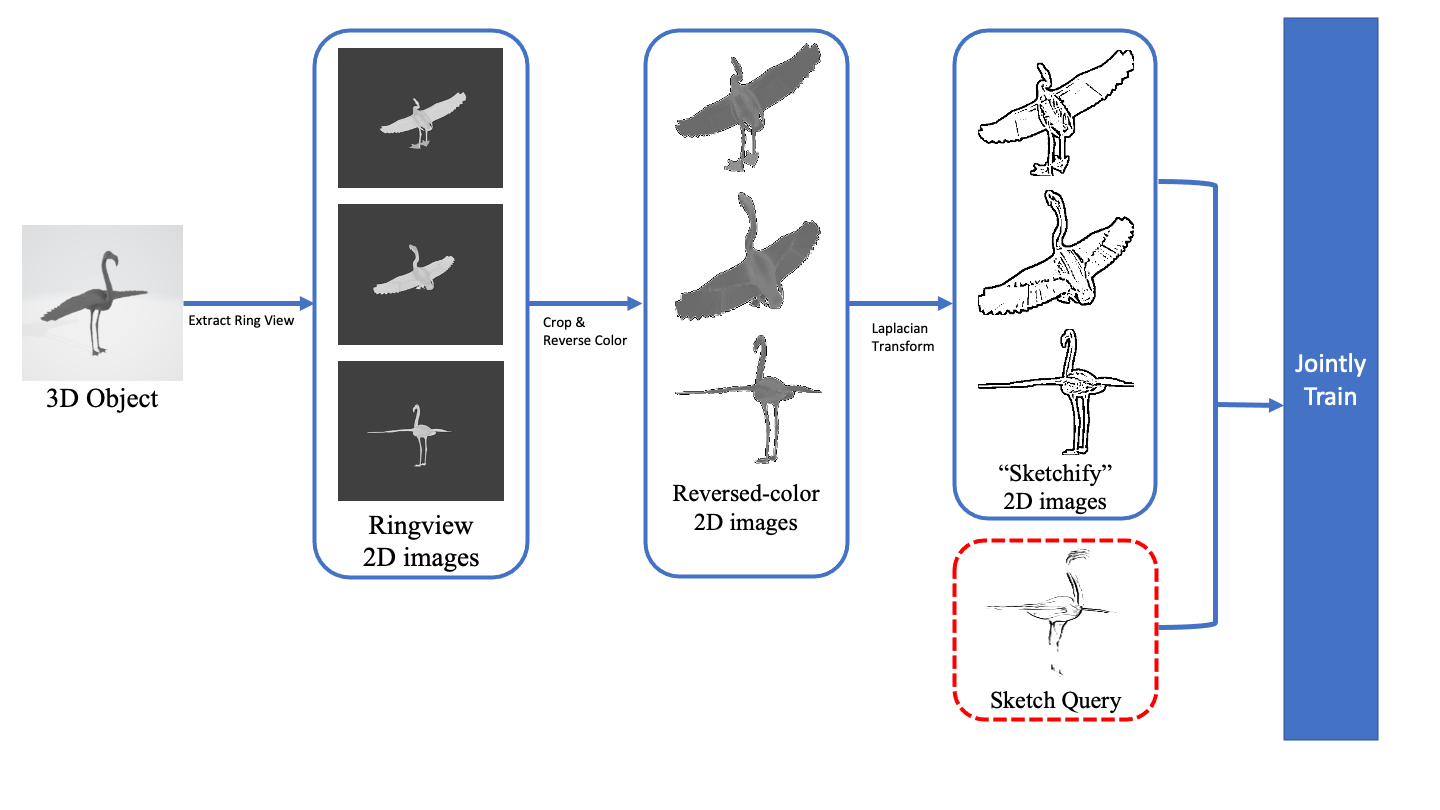}
    \caption{The proposed data pre-processing pipeline}
    \label{fig:pipeline}
\end{figure}

\subsubsection{Data Pre-processing}

Figure~\ref{fig:pipeline} demonstrates an overview of the proposed data pre-processing pipeline. In general, it can be divided into three steps: Ringview extraction, color reversal, and sketchify.

\begin{figure*}[t!]
    \centering
    \includegraphics[width=\linewidth]{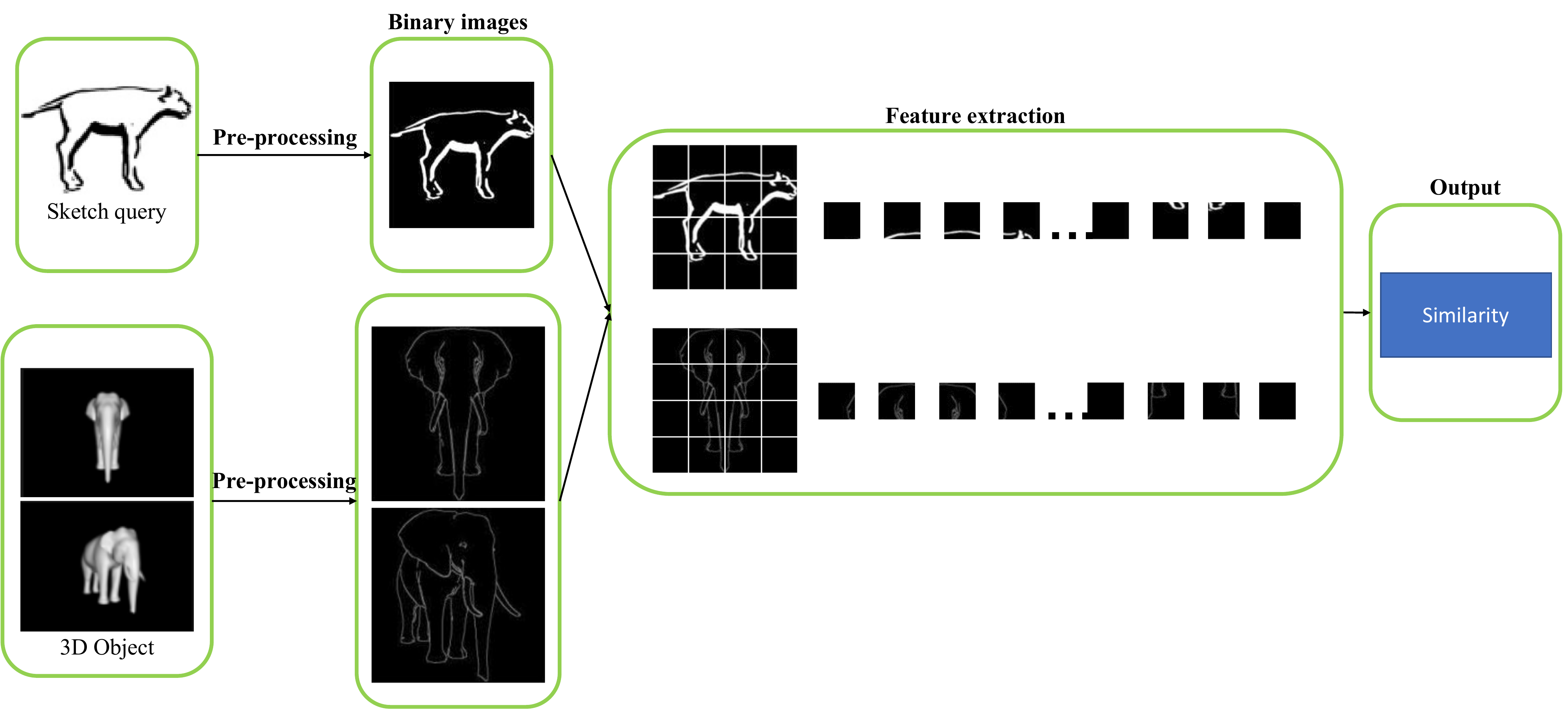}
    \caption{Proposed framework of DH team.}
    \label{fig:overview_dh}
\end{figure*}

\textbf{Ringview extraction.} Extracting multiple views of an object can be highly advantageous for various applications, including 3D object retrieval. They extract multiple views from 7 rings with 12 views on each ring, like Fig.~\ref{fig:multiviewrepresentation}. By providing different perspectives of the 3D object, these multiple views can extract more robust and detailed features to train models with greater accuracy for 3D object retrieval. Thus, ringview processing is a valuable technique that can improve the accuracy of models for 3D object retrieval.

\textbf{Color reversal.} It is also crucial to consider the background of the images when matching 3D objects with the querying sketches. Typically, these sketches usually have a white background, while the multiple 2D images obtained from the ringview extraction step have a grey background. To solve this problem, they merely flip the color of the ringview image, making the backdrop translucent to match the background of the target sketch queries. Therefore, the resulting images better resemble the sketch queries and support the further "sketchify" procedure.

\textbf{Sketchify.} Laplacian Transform is utilized to produce images that are more similar to sketches. Particularly, the Laplacian Transform, as a linear operator, is applied to the gray-scale image to generate a second-order derivative image that enhances the edges and transitions of the image~\cite{Nealen-SIGGRAPH2005}. The operator produces a sketch-like version of an image by thresholding the Laplacian image to obtain a binary edge map, which is used to synthesize a sketch-like representation of the image. This transformation enables more efficient comparison of the views with the sketch query as the critical matching features become more pronounced, facilitating accurate matching.


\subsection{DH Team}
\label{sec:team_dh}

\subsubsection{Proposed Method}

They propose a simple method to measure the distance between 2 images with fast execution speed due to the small number of computations. Based on the observation that 3D objects with the same shape will have the same distribution of pixels in 2D projection space, they propose dividing the image into small parts and then comparing each area to measure the similarity between the two images. Fig.~\ref{fig:overview_dh} shows the proposed framework, including four main modules: 2D sketch processing, 3D model processing, image feature extraction, and feature matching.

\textbf{2D sketch processing.} They crop images intending to keep only the part containing the animals and remove the background. Since sketches in the dataset only contain the animal and no other extraneous details in the background, they define the animal's bounding box as follows: First, sketches are converted to binary images and then inverted. Then they find the top and bottom image lines with the value 255. These two lines correspond to the top and bottom edges of the bounding box. The same method is applied to the left and right edges. After that, they crop the part containing the animal into a square with the size of a maximum bounding-box height and width and then resize the image to $224 \times 224$.

\textbf{3D model processing. } They first rotate the 3D objects using the Open3D library and capture the 3D model from 21 perspectives. After that, view images are cropped similarly to 2D sketches.

\textbf{Image feature extraction.} The images are divided into $4 \times 4$ squares, and then the ratio of total pixel values in each square to total pixel values of the whole image is considered at the score of the square. At the end of this step, a 16-dimensional feature vector represents an image, including the sketch and view images.

\textbf{Feature matching.} Given a sketch query image $Q$ and a 3D model $R$, features are extracted to obtain $f^Q$ representing the sketch image and $f^R_i, i=1..21$ representing the 21 view images corresponding to the 3D model. The similarity score between the sketch query $Q$ and the 3D model $R$ is defined as follows:
\begin{equation}
    D(Q, R) = min_{i}||f^Q - f^R_i||_2.
\end{equation}


\section{Results and Discussions}
\label{sec:results}

\begin{table*}[!t]
    \caption{Leaderboard results of SketchANIMAR competition. Best run results on the public test.}
    \label{tab:sketch-public}
    \centering
    \begin{tabular}{|l|c|c|c|c|c|c|c|}
    \hline
    \multicolumn{1}{|c|}{\textbf{Team}} &
      \multicolumn{1}{c|}{\textbf{NN}} &
      \multicolumn{1}{c|}{\textbf{P@10}} &
      \multicolumn{1}{c|}{\textbf{NDCG}} &
      \multicolumn{1}{c|}{\textbf{mAP}} &
      \multicolumn{1}{c|}{\textbf{FT}} &
      \multicolumn{1}{c|}{\textbf{ST}} &
      \multicolumn{1}{c|}{\textbf{FR}} \\ \hline
    \rowcolor{lightgray} TikTorch              & \textbf{0.533 (1)}   & \textbf{0.280 (1)} & \textbf{0.708 (1)} & \textbf{0.570 (1)} &  0.192 (4) & 0.333 (4) & \textbf{0.0102 (1)} \\ 
    V1olet                & 0.467 (2)   & 0.213 (2) & 0.613 (2) & 0.411 (2)  & 0.317 (2) & 0.492 (2) & 0.0111 (2) \\ 
    THP                   & 0.433 (3)   & 0.207 (3) & 0.601 (3) & 0.399 (3)  & 0.300 (3) & 0.450 (3) & 0.0112 (3) \\ 
    Etinifni              & 0.200 (4)   & 0.147 (4) & 0.489 (4) & 0.303 (4)  & \textbf{0.475 (1)} & \textbf{0.650 (1)} & 0.0121 (4) \\ 
    DH                    & 0.100 (5)   & 0.080 (5) & 0.361 (5) & 0.140 (5)  & 0.133 (5) & 0.192 (5) & 0.0130 (5) \\ 
    \hline
    \end{tabular}%
\end{table*}

\begin{table*}[!t]
    \caption{Leaderboard results of SketchANIMAR competition. Best run results on the private test.}
    \label{tab:sketch-private}
    \centering
    \begin{tabular}{|l|c|c|c|c|c|c|c|}
    \hline
    \multicolumn{1}{|c|}{\textbf{Team}} &
      \multicolumn{1}{c|}{\textbf{NN}} &
      \multicolumn{1}{c|}{\textbf{P@10}} &
      \multicolumn{1}{c|}{\textbf{NDCG}} &
      \multicolumn{1}{c|}{\textbf{mAP}} &
      \multicolumn{1}{c|}{\textbf{FT}} &
      \multicolumn{1}{c|}{\textbf{ST}} &
      \multicolumn{1}{c|}{\textbf{FR}} \\ \hline
    \rowcolor{lightgray} TikTorch              & 0.470 (2) & \textbf{0.255 (1)} & \textbf{0.669 (1)} & \textbf{0.522 (1)} & \textbf{0.424 (1)} & \textbf{0.583 (1)} & \textbf{0.0105 (1)} \\ 
    V1olet                & \textbf{0.500 (1)} & 0.232 (2) & 0.640 (2) & 0.453 (2) & 0.379 (2) & 0.534 (2) & 0.0109 (2) \\ 
    THP                   & 0.409 (3) & 0.226 (3) & 0.608 (3) & 0.421 (3) & 0.333 (3) & 0.504 (3) & 0.0110 (3) \\ 
    Etinifni              & 0.136 (4) & 0.158 (4) & 0.473 (4) & 0.274 (4) & 0.174 (4) & 0.345 (4) & 0.0119 (4) \\ 
    DH                    & 0.136 (4) & 0.088 (5) & 0.372 (5) & 0.158 (5) & 0.133 (5) & 0.205 (5) & 0.0129 (5) \\ 
    \hline
    \end{tabular}%
    \vspace{-3mm}
\end{table*}

In our SketchANIMAR track, the submitted runs are evaluated on two subsets: the public and private tests. The private test comprised 66 sketch queries, resulting in 265 query-model pairs. To ensure fairness and prevent potential cheating, approximately half of the private test (30 sketch queries) was randomly selected and designated as the public test subset. The leaderboard for the private test was revealed after the challenge concluded.

Tables~\ref{tab:sketch-public} and~\ref{tab:sketch-private} display the leaderboard outcomes for the public and the private test, reporting only the best-performing runs submitted by each team. However, to ensure a fair comparison, our analysis focuses solely on the results from the private test, which evaluated all the submitted sketch queries.

As seen in Table~\ref{tab:sketch-private}, the TikTorch and V1olet team's presented methods repeatedly stood out as the top effective strategies. \highlightx{On $6$ out of $7$ performance metrics (P@10, NDCG, mAP, FT, ST, and FR), TikTorch outperformed rival teams by a wide margin.} In terms of NN metric, the V1olet team secured the top 1 position, \highlightx{indicating that this team focuses on the best search instead of neighboring search}. Meanwhile, THP took up the third position on public and private leaderboards. Most teams achieving high performances apply the view-base approach, which analyzes shape features from 2D view projections. For the public test, similar findings are also shown in Table~\ref{tab:sketch-public}, \highlightx{excepting that the top-1 team (TikTorch) only takes the fourth position on FT and ST metrics}. The first two teams, TikTorck and V1olet, achieve the top two ranks. In contrast, the performance of the other teams (\ie, THP, Etinifni, and DH) is constant across test sets, both private and public. \highlightx{Summary, the findings show the challenges of our ANIMAR dataset when participants did not achieve excellent performance (the best NN result is only around 0.5, and P@10 results are smaller than 0.3). It also suggests that there is room to improve the performance of this research direction.}

\begin{figure}[!t]
    \centering
    \includegraphics[width=\columnwidth]{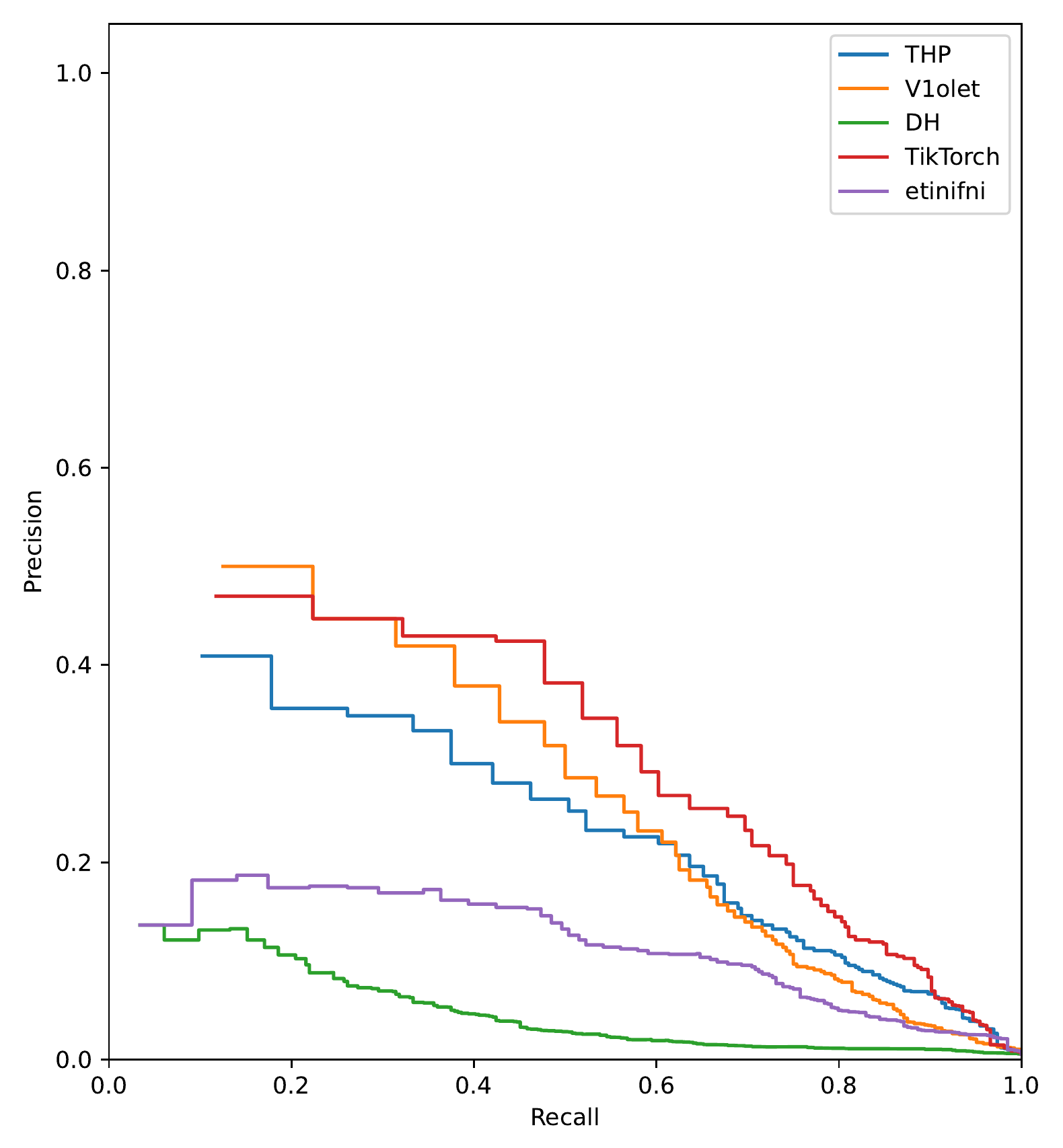}
    \caption{The visualization of precision-recall curves of submissions on the private test of teams. It can be seen that the TikTorch team achieves the best average performance with the highest area under the curve, while V1olet obtains the highest precision at low recall (recall $\le 0.2$) among the eight teams. These insights are compatible with results in Table~\ref{tab:sketch-private}, TikTorch and V1olet secure the top 1 position regarding mAP and NN, respectively.}
    \label{fig:pr_curves}
\end{figure}

Figure~\ref{fig:pr_curves} illustrates the precision-recall curves of submissions on the private test of teams. It is clear that among the teams, the TikTorch team has the best average performance with the biggest area under the curve, while V1olet has the maximum precision at low recall (recall $\le 0.2$). \highlightx{The main difference between TikTorch's and V1olet's methods is that TikTorch follows a contrastive learning approach while V1olet leverages a classification-based one with a softmax layer. The softmax score just implies whether an object and a sketched query belong to the same category and can not indicate much similarity between them. In general, V1olet's method works with objects in the same category as sketched queries and is less robust to negative samples than the contrastive learning approach.} These conclusions are consistent with Table~\ref{tab:sketch-private} 's findings, which show that TikTorch and V1olet rank first in terms of mAP and NN. Furthermore, the precision-recall curves of three teams, TikTorch, V1olet, and THP, share a similar shape, which shows the effectiveness at low recall threshold (recall $\le 0.4$) but drops significantly at high one (recall $\ge 0.8$).

In conclusion, view-based learning methods have proven effective in achieving high performance. The difficulty of feature extraction models~\cite{Qi-CVPR2017, Muzahid-JAS2020} can be attributed to the high point cloud density of the 3D objects in the ANIMAR dataset. It is important to remember that these models often randomly sample a certain number of \highlightx{pointclouds} (\eg, 1024). Contrarily, by utilizing the semantic data and representation of 3D objects, the use of view pictures obtained by moving the trajectory camera, as demonstrated in Fig.~\ref{fig:multiviewrepresentation}, enhances feature learning. This further demonstrates the effectiveness of the view-based learning strategy for retrieving 3D objects.

\section{Conclusion}
\label{sec:conclusion}

This paper introduces a novel track for sketch-based retrieval of fine-grained 3D animal models along with a newly constructed ANIMAR dataset. Our SHREC 2023 challenge track is designed to simulate real-life scenarios and requires participants to retrieve 3D animal models based on complex and detailed sketches. The challenge received submissions from eight teams; however, \highlightx{the evaluated results in this paper include methods from five of the eight teams.} These submissions resulted in a total of 204 runs with different approaches. The evaluated results of this track were satisfactory but also revealed the difficulties of the task at hand.

In future research, we aim to expand the dataset by collecting a more diverse range of 3D animal models that encompass a wider variety of species, environmental contexts, and postures. This can enhance the generalization capability of potential solutions and improve performance on unseen 3D animal models. Additionally, we intend to generate synthetic data and texture to augment 3D animal models with different postures, backgrounds, and patterns to train more effective and robust representation models. We believe that by exploring these research avenues, we can advance the state-of-the-art in 3D object retrieval.

\section*{CRediT authorship contribution statement}


\textbf{Trung-Nghia Le}: Conceptualization, Writing – review $\&$ editing, Project administration, Supervision. 
\textbf{Tam V. Nguyen}: Conceptualization,  Writing – review $\&$ editing. 
\textbf{Minh-Quan Le}: Software, Writing – review $\&$ editing. 
\textbf{Trong-Thuan Nguyen}: Software, Writing – review $\&$ editing.
\textbf{Viet-Tham Huynh}: Data curation. 
\textbf{Trong-Le Do}: Software, Investigation.
\textbf{Khanh-Duy Le}: Visualization. 
\textbf{Mai-Khiem Tran}: Data curation. 
\textbf{Nhat Hoang-Xuan}: Software. 
\textbf{Thang-Long Nguyen-Ho}: Software. 
\textbf{Vinh-Tiep Nguyen}: Conceptualization. 
\textbf{Nhat-Quynh Le-Pham}: Methodology, Writing – original draft. 
\textbf{Huu-Phuc Pham}: Methodology, Writing – original draft. 
\textbf{Trong-Vu Hoang}: Methodology, Writing – original draft. 
\textbf{Quang-Binh Nguyen}: Methodology, Writing – original draft. 
\textbf{Hai-Dang Nguyen}: Methodology, Writing – original draft. 
\textbf{Trong-Hieu Nguyen-Mau}: Methodology, Writing – original draft. 
\textbf{Tuan-Luc Huynh}: Methodology, Writing – original draft. 
\textbf{Thanh-Danh Le}: Methodology, Writing – original draft. 
\textbf{Ngoc-Linh Nguyen-Ha}: Methodology, Writing – original draft. 
\textbf{Tuong-Vy Truong-Thuy}: Methodology, Writing – original draft. 
\textbf{Truong Hoai Phong}: Methodology, Writing – original draft. 
\textbf{Tuong-Nghiem Diep}: Methodology, Writing – original draft.
\textbf{Khanh-Duy Ho}: Methodology, Writing – original draft.
\textbf{Xuan-Hieu Nguyen}: Methodology, Writing – original draft.
\textbf{Thien-Phuc Tran}: Methodology, Writing – original draft.
\textbf{Tuan-Anh Yang}: Methodology, Writing – original draft.
\textbf{Kim-Phat Tran}: Methodology, Writing – original draft.
\textbf{Nhu-Vinh Hoang}: Methodology, Writing – original draft.
\textbf{Minh-Quang Nguyen}: Methodology, Writing – original draft. 
\textbf{Hoai-Danh Vo}: Methodology, Writing – original draft. 
\textbf{Minh-Hoa Doan}: Methodology, Writing – original draft. 
\textbf{Akihiro Sugimoto}: Conceptualization.
\textbf{Minh-Triet Tran}: Conceptualization, Supervision, Funding acquisition, Writing - Review \& Editing.

\section*{Data availability}

\highlight{After the challenge concluded, the dataset has been made publicly available for academic purposes.}

\section*{Declaration of competing interest}

The authors declare that they have no known competing financial interests or personal relationships that could have appeared to influence the work reported in this paper.

\section*{Acknowledgments}


This work was funded by the Vingroup Innovation Foundation (VINIF.2019.DA19) and National Science Foundation Grant (NSF\#2025234). 

\bibliographystyle{cag-num-names}
\bibliography{short_bibtex}

\begin{thebibliography}{48}
\providecommand{\natexlab}[1]{#1}
\providecommand{\url}[1]{\texttt{#1}}
\providecommand{\href}[2]{#2}
\providecommand{\path}[1]{#1}
\providecommand{\eprint}[1]{\href{http://arxiv.org/abs/#1}{\path{#1}}}
\providecommand{\DOIprefix}{doi:}
\providecommand{\ArXivprefix}{arXiv:}
\providecommand{\URLprefix}{URL: }
\providecommand{\Pubmedprefix}{pmid:}
\providecommand{\doi}[1]{\href{http://dx.doi.org/#1}{\path{#1}}}
\providecommand{\Pubmed}[1]{\href{pmid:#1}{\path{#1}}}
\providecommand{\BIBand}{and}
\providecommand{\bibinfo}[2]{#2}
\ifx\xfnm\undefined \def\xfnm[#1]{\unskip,\space#1}\fi
\bibitem[{Stotko et~al.(2019)Stotko, Krumpen, Hullin, Weinmann and
  Klein}]{stotko2019slamcast}
\bibinfo{author}{Stotko\xfnm[ P]}, \bibinfo{author}{Krumpen\xfnm[ S]},
  \bibinfo{author}{Hullin\xfnm[ MB]}, \bibinfo{author}{Weinmann\xfnm[ M]},
  \bibinfo{author}{Klein\xfnm[ R]}.
\newblock \bibinfo{title}{Slamcast: Large-scale, real-time {3D} reconstruction
  and streaming for immersive multi-client live telepresence}.
\newblock \bibinfo{journal}{IEEE transactions on visualization and computer
  graphics}
  \bibinfo{year}{2019};\bibinfo{volume}{25}(\bibinfo{number}{5}):\bibinfo{pages}{2102--2112}.
\bibitem[{Liu and Kofman(2019)}]{liu2019real}
\bibinfo{author}{Liu\xfnm[ X]}, \bibinfo{author}{Kofman\xfnm[ J]}.
\newblock \bibinfo{title}{Real-time {3D} surface-shape measurement using
  background-modulated modified fourier transform profilometry with
  geometry-constraint}.
\newblock \bibinfo{journal}{Optics and Lasers in Engineering}
  \bibinfo{year}{2019};\bibinfo{volume}{115}:\bibinfo{pages}{217--224}.
\bibitem[{Wang et~al.(2020)Wang, Mueller, Bernard, Sorli, Sotnychenko, Qian
  et~al.}]{wang2020rgb2hands}
\bibinfo{author}{Wang\xfnm[ J]}, \bibinfo{author}{Mueller\xfnm[ F]},
  \bibinfo{author}{Bernard\xfnm[ F]}, \bibinfo{author}{Sorli\xfnm[ S]},
  \bibinfo{author}{Sotnychenko\xfnm[ O]}, \bibinfo{author}{Qian\xfnm[ N]},
  et~al.
\newblock \bibinfo{title}{{RGB}2hands: real-time tracking of {3D} hand
  interactions from monocular {RGB} video}.
\newblock \bibinfo{journal}{ACM Transactions on Graphics (ToG)}
  \bibinfo{year}{2020};\bibinfo{volume}{39}(\bibinfo{number}{6}):\bibinfo{pages}{1--16}.
\bibitem[{Guo et~al.(2022)Guo, Peng, Lin, Wang, Zhang, Bao
  et~al.}]{guo2022neural}
\bibinfo{author}{Guo\xfnm[ H]}, \bibinfo{author}{Peng\xfnm[ S]},
  \bibinfo{author}{Lin\xfnm[ H]}, \bibinfo{author}{Wang\xfnm[ Q]},
  \bibinfo{author}{Zhang\xfnm[ G]}, \bibinfo{author}{Bao\xfnm[ H]}, et~al.
\newblock \bibinfo{title}{Neural {3D} scene reconstruction with the
  manhattan-world assumption}.
\newblock In: \bibinfo{booktitle}{Proceedings of the IEEE/CVF Conference on
  Computer Vision and Pattern Recognition}. \bibinfo{year}{2022}, p.
  \bibinfo{pages}{5511--5520}.
\bibitem[{Yookwan et~al.(2022)Yookwan, Chinnasarn, So-In and
  Horkaew}]{yookwan2022multimodal}
\bibinfo{author}{Yookwan\xfnm[ W]}, \bibinfo{author}{Chinnasarn\xfnm[ K]},
  \bibinfo{author}{So-In\xfnm[ C]}, \bibinfo{author}{Horkaew\xfnm[ P]}.
\newblock \bibinfo{title}{Multimodal fusion of deeply inferred point clouds for
  {3D} scene reconstruction using cross-entropy icp}.
\newblock \bibinfo{journal}{IEEE Access}
  \bibinfo{year}{2022};\bibinfo{volume}{10}:\bibinfo{pages}{77123--77136}.
\bibitem[{Li et~al.(2022)Li, Gao, Wu, Liu and Shen}]{li2022high}
\bibinfo{author}{Li\xfnm[ J]}, \bibinfo{author}{Gao\xfnm[ W]},
  \bibinfo{author}{Wu\xfnm[ Y]}, \bibinfo{author}{Liu\xfnm[ Y]},
  \bibinfo{author}{Shen\xfnm[ Y]}.
\newblock \bibinfo{title}{High-quality indoor scene {3D} reconstruction with
  {RGB-D} cameras: A brief review}.
\newblock \bibinfo{journal}{Computational Visual Media}
  \bibinfo{year}{2022};\bibinfo{volume}{8}(\bibinfo{number}{3}):\bibinfo{pages}{369--393}.
\bibitem[{G{\"u}meli et~al.(2022)G{\"u}meli, Dai and
  Nie{\ss}ner}]{gumeli2022roca}
\bibinfo{author}{G{\"u}meli\xfnm[ C]}, \bibinfo{author}{Dai\xfnm[ A]},
  \bibinfo{author}{Nie{\ss}ner\xfnm[ M]}.
\newblock \bibinfo{title}{Roca: robust {CAD} model retrieval and alignment from
  a single image}.
\newblock In: \bibinfo{booktitle}{Proceedings of the IEEE/CVF Conference on
  Computer Vision and Pattern Recognition}. \bibinfo{year}{2022}, p.
  \bibinfo{pages}{4022--4031}.
\bibitem[{Manda et~al.(2022)Manda, Kendre, Dey and
  Muthuganapathy}]{manda2022sketchcleannet}
\bibinfo{author}{Manda\xfnm[ B]}, \bibinfo{author}{Kendre\xfnm[ PP]},
  \bibinfo{author}{Dey\xfnm[ S]}, \bibinfo{author}{Muthuganapathy\xfnm[ R]}.
\newblock \bibinfo{title}{Sketchcleannet—a deep learning approach to the
  enhancement and correction of query sketches for a {3D} {CAD} model retrieval
  system}.
\newblock \bibinfo{journal}{Computers and Graphics}
  \bibinfo{year}{2022};\bibinfo{volume}{107}:\bibinfo{pages}{73--83}.
\bibitem[{Salihu and Steinbach(2023)}]{salihu2023sgpcr}
\bibinfo{author}{Salihu\xfnm[ D]}, \bibinfo{author}{Steinbach\xfnm[ E]}.
\newblock \bibinfo{title}{{SGPCR}: Spherical gaussian point cloud
  representation and its application to object registration and retrieval}.
\newblock In: \bibinfo{booktitle}{Proceedings of the IEEE/CVF Winter Conference
  on Applications of Computer Vision}. \bibinfo{year}{2023}, p.
  \bibinfo{pages}{572--581}.
\bibitem[{Koca et~al.(2019)Koca, {\c{C}}ubuk{\c{c}}u and
  Y{\"u}zge{\c{c}}}]{koca2019augmented}
\bibinfo{author}{Koca\xfnm[ BA]}, \bibinfo{author}{{\c{C}}ubuk{\c{c}}u\xfnm[
  B]}, \bibinfo{author}{Y{\"u}zge{\c{c}}\xfnm[ U]}.
\newblock \bibinfo{title}{Augmented reality application for preschool children
  with unity {3D} platform}.
\newblock In: \bibinfo{booktitle}{2019 3rd International Symposium on
  Multidisciplinary Studies and Innovative Technologies (ISMSIT)}.
  \bibinfo{organization}{IEEE}; \bibinfo{year}{2019}, p. \bibinfo{pages}{1--4}.
\bibitem[{Guo et~al.(2023)Guo, Jiang, Chen, Song and
  Hilliges}]{guo2023vid2avatar}
\bibinfo{author}{Guo\xfnm[ C]}, \bibinfo{author}{Jiang\xfnm[ T]},
  \bibinfo{author}{Chen\xfnm[ X]}, \bibinfo{author}{Song\xfnm[ J]},
  \bibinfo{author}{Hilliges\xfnm[ O]}.
\newblock \bibinfo{title}{Vid2avatar: {3D} avatar reconstruction from videos in
  the wild via self-supervised scene decomposition}.
\newblock \bibinfo{journal}{arXiv preprint arXiv:230211566}
  \bibinfo{year}{2023};.
\bibitem[{Li et~al.(2012)Li, Schreck, Godil, Alexa, Boubekeur, Bustos
  et~al.}]{Li-SHREC2012}
\bibinfo{author}{Li\xfnm[ B]}, \bibinfo{author}{Schreck\xfnm[ T]},
  \bibinfo{author}{Godil\xfnm[ A]}, \bibinfo{author}{Alexa\xfnm[ M]},
  \bibinfo{author}{Boubekeur\xfnm[ T]}, \bibinfo{author}{Bustos\xfnm[ B]},
  et~al.
\newblock \bibinfo{title}{{SHREC}'12 track: Sketch-based {3D} shape retrieval}.
\newblock In: \bibinfo{booktitle}{{3D}OR@ Eurographics}. \bibinfo{year}{2012},
  p. \bibinfo{pages}{109--118}.
\bibitem[{Li et~al.(2013)Li, Lu, Godil, Schreck, Aono, Johan
  et~al.}]{Li-SHREC2013}
\bibinfo{author}{Li\xfnm[ B]}, \bibinfo{author}{Lu\xfnm[ Y]},
  \bibinfo{author}{Godil\xfnm[ A]}, \bibinfo{author}{Schreck\xfnm[ T]},
  \bibinfo{author}{Aono\xfnm[ M]}, \bibinfo{author}{Johan\xfnm[ H]}, et~al.
\newblock \bibinfo{title}{SHREC’13 track: Large scale sketch-based {3D} shape
  retrieval}.
\newblock \bibinfo{year}{2013}.
\bibitem[{Li et~al.(2014)Li, Lu, Li, Godil, Schreck, Aono
  et~al.}]{Li-SHREC2014}
\bibinfo{author}{Li\xfnm[ B]}, \bibinfo{author}{Lu\xfnm[ Y]},
  \bibinfo{author}{Li\xfnm[ C]}, \bibinfo{author}{Godil\xfnm[ A]},
  \bibinfo{author}{Schreck\xfnm[ T]}, \bibinfo{author}{Aono\xfnm[ M]}, et~al.
\newblock \bibinfo{title}{Shrec’14 track: Extended large scale sketch-based
  {3D} shape retrieval}.
\newblock In: \bibinfo{booktitle}{Eurographics workshop on {3D} object
  retrieval}; vol. \bibinfo{volume}{2014}. \bibinfo{year}{2014}, p.
  \bibinfo{pages}{121--130}.
\bibitem[{Yuan et~al.(2018)Yuan, Li, Lu, Bai, Bai, Bui
  et~al.}]{Juefei-SHREC2018}
\bibinfo{author}{Yuan\xfnm[ J]}, \bibinfo{author}{Li\xfnm[ B]},
  \bibinfo{author}{Lu\xfnm[ Y]}, \bibinfo{author}{Bai\xfnm[ S]},
  \bibinfo{author}{Bai\xfnm[ X]}, \bibinfo{author}{Bui\xfnm[ NM]}, et~al.
\newblock \bibinfo{title}{Shrec’18 track: 2d scene sketch-based {3D} scene
  retrieval}.
\newblock \bibinfo{journal}{Eurographics Workshop on {3D} Object Retrieval}
  \bibinfo{year}{2018};\bibinfo{volume}{18}:\bibinfo{pages}{70}.
\bibitem[{Yuan et~al.(2019)Yuan, Abdul-Rashid, Li, Lu, Schreck, Bui
  et~al.}]{Juefei-SHREC2019}
\bibinfo{author}{Yuan\xfnm[ J]}, \bibinfo{author}{Abdul-Rashid\xfnm[ H]},
  \bibinfo{author}{Li\xfnm[ B]}, \bibinfo{author}{Lu\xfnm[ Y]},
  \bibinfo{author}{Schreck\xfnm[ T]}, \bibinfo{author}{Bui\xfnm[ NM]}, et~al.
\newblock \bibinfo{title}{Shrec’19 track: Extended 2d scene sketch-based {3D}
  scene retrieval}.
\newblock \bibinfo{journal}{Eurographics Workshop on {3D} Object Retrieval}
  \bibinfo{year}{2019};\bibinfo{volume}{18}:\bibinfo{pages}{70}.
\bibitem[{Qin et~al.(2022)Qin, Yuan, Chen, {Ben Amor}, Fang, Hoang-Xuan
  et~al.}]{Qin-SHREC2022}
\bibinfo{author}{Qin\xfnm[ J]}, \bibinfo{author}{Yuan\xfnm[ S]},
  \bibinfo{author}{Chen\xfnm[ J]}, \bibinfo{author}{{Ben Amor}\xfnm[ B]},
  \bibinfo{author}{Fang\xfnm[ Y]}, \bibinfo{author}{Hoang-Xuan\xfnm[ N]},
  et~al.
\newblock \bibinfo{title}{Shrec’22 track: Sketch-based {3D} shape retrieval
  in the wild}.
\newblock \bibinfo{journal}{Computers and Graphics} \bibinfo{year}{2022};.
\bibitem[{Abdul-Rashid et~al.(2018)Abdul-Rashid, Yuan, Li, Lu, Bai, Bai
  et~al.}]{Hameed-SHREC2018}
\bibinfo{author}{Abdul-Rashid\xfnm[ H]}, \bibinfo{author}{Yuan\xfnm[ J]},
  \bibinfo{author}{Li\xfnm[ B]}, \bibinfo{author}{Lu\xfnm[ Y]},
  \bibinfo{author}{Bai\xfnm[ S]}, \bibinfo{author}{Bai\xfnm[ X]}, et~al.
\newblock \bibinfo{title}{{2D Image-Based {3D} Scene Retrieval}}.
\newblock In: \bibinfo{editor}{Telea\xfnm[ A]},
  \bibinfo{editor}{Theoharis\xfnm[ T]}, \bibinfo{editor}{Veltkamp\xfnm[ R]},
  editors. \bibinfo{booktitle}{Eurographics Workshop on {3D} Object Retrieval}.
  \bibinfo{year}{2018},.
\bibitem[{Abdul-Rashid et~al.(2019)Abdul-Rashid, Yuan, Li, Lu, Schreck, Bui
  et~al.}]{Hameed-SHREC2019}
\bibinfo{author}{Abdul-Rashid\xfnm[ H]}, \bibinfo{author}{Yuan\xfnm[ J]},
  \bibinfo{author}{Li\xfnm[ B]}, \bibinfo{author}{Lu\xfnm[ Y]},
  \bibinfo{author}{Schreck\xfnm[ T]}, \bibinfo{author}{Bui\xfnm[ NM]}, et~al.
\newblock \bibinfo{title}{{SHREC}’19 track: Extended 2d scene image-based
  {3D} scene retrieval}.
\newblock \bibinfo{journal}{Eurographics Workshop on {3D} Object Retrieval}
  \bibinfo{year}{2019};\bibinfo{volume}{700}:\bibinfo{pages}{70}.
\bibitem[{Li et~al.(2019)Li, Liu, Nie, Song, Li, Wang et~al.}]{Li-SHREC2019}
\bibinfo{author}{Li\xfnm[ W]}, \bibinfo{author}{Liu\xfnm[ A]},
  \bibinfo{author}{Nie\xfnm[ W]}, \bibinfo{author}{Song\xfnm[ D]},
  \bibinfo{author}{Li\xfnm[ Y]}, \bibinfo{author}{Wang\xfnm[ W]}, et~al.
\newblock \bibinfo{title}{{SHREC} 2019-monocular image based {3D} model
  retrieval}.
\newblock In: \bibinfo{booktitle}{Eurographics Workshop {3D} Object Retrieval}.
  \bibinfo{year}{2019}, p. \bibinfo{pages}{1--8}.
\bibitem[{Li et~al.(2020)Li, Song, Liu, Nie, Zhang, Zhao et~al.}]{Li-SHREC2020}
\bibinfo{author}{Li\xfnm[ W]}, \bibinfo{author}{Song\xfnm[ D]},
  \bibinfo{author}{Liu\xfnm[ A]}, \bibinfo{author}{Nie\xfnm[ W]},
  \bibinfo{author}{Zhang\xfnm[ T]}, \bibinfo{author}{Zhao\xfnm[ X]}, et~al.
\newblock \bibinfo{title}{{SHREC} 2020 track: extended monocular image based 3d
  model retrieval}.
\newblock In: \bibinfo{booktitle}{Eurographics Workshop {3D} Object Retrieval}.
  \bibinfo{year}{2020},.
\bibitem[{Feng et~al.(2022)Feng, Gao, Zhao, Guo, Bagewadi, Bui
  et~al.}]{Feng-SHREC2022}
\bibinfo{author}{Feng\xfnm[ Y]}, \bibinfo{author}{Gao\xfnm[ Y]},
  \bibinfo{author}{Zhao\xfnm[ X]}, \bibinfo{author}{Guo\xfnm[ Y]},
  \bibinfo{author}{Bagewadi\xfnm[ N]}, \bibinfo{author}{Bui\xfnm[ NT]}, et~al.
\newblock \bibinfo{title}{{SHREC}’22 track: Open-set {3D} object retrieval}.
\newblock \bibinfo{journal}{Computers \& Graphics}
  \bibinfo{year}{2022};\bibinfo{volume}{107}:\bibinfo{pages}{231--240}.
\bibitem[{Furuya and Ohbuchi(2016)}]{Furuya-BMVC2016}
\bibinfo{author}{Furuya\xfnm[ T]}, \bibinfo{author}{Ohbuchi\xfnm[ R]}.
\newblock \bibinfo{title}{Deep aggregation of local {3D} geometric features for
  {3D} model retrieval}.
\newblock In: \bibinfo{booktitle}{Proceedings of the British Machine Vision
  Conference (BMVC)}. \bibinfo{year}{2016},.
\bibitem[{Wang et~al.(2017)Wang, Liu, Guo, Sun and Tong}]{Wang-TransGraph2017}
\bibinfo{author}{Wang\xfnm[ PS]}, \bibinfo{author}{Liu\xfnm[ Y]},
  \bibinfo{author}{Guo\xfnm[ YX]}, \bibinfo{author}{Sun\xfnm[ CY]},
  \bibinfo{author}{Tong\xfnm[ X]}.
\newblock \bibinfo{title}{O-{CNN}: Octree-based convolutional neural networks
  for {3D} shape analysis}.
\newblock \bibinfo{journal}{ACM Trans Graph} \bibinfo{year}{2017};.
\bibitem[{Wang et~al.(2015)Wang, Shen and Porikli}]{Wang-CVPR2015}
\bibinfo{author}{Wang\xfnm[ W]}, \bibinfo{author}{Shen\xfnm[ J]},
  \bibinfo{author}{Porikli\xfnm[ F]}.
\newblock \bibinfo{title}{Saliency-aware geodesic video object segmentation}.
\newblock In: \bibinfo{booktitle}{CVPR}. \bibinfo{year}{2015}, p.
  \bibinfo{pages}{3395--3402}.
\bibitem[{Xie et~al.(2017)Xie, Girshick, Doll{\'a}r, Tu and He}]{Xie-CVPR2017}
\bibinfo{author}{Xie\xfnm[ S]}, \bibinfo{author}{Girshick\xfnm[ R]},
  \bibinfo{author}{Doll{\'a}r\xfnm[ P]}, \bibinfo{author}{Tu\xfnm[ Z]},
  \bibinfo{author}{He\xfnm[ K]}.
\newblock \bibinfo{title}{Aggregated residual transformations for deep neural
  networks}.
\newblock In: \bibinfo{booktitle}{CVPR}. \bibinfo{year}{2017}, p.
  \bibinfo{pages}{1492--1500}.
\bibitem[{Su et~al.(2015)Su, Maji, Kalogerakis and
  Learned{-}Miller}]{Su-ICCV2015}
\bibinfo{author}{Su\xfnm[ H]}, \bibinfo{author}{Maji\xfnm[ S]},
  \bibinfo{author}{Kalogerakis\xfnm[ E]},
  \bibinfo{author}{Learned{-}Miller\xfnm[ EG]}.
\newblock \bibinfo{title}{Multi-view convolutional neural networks for {3D}
  shape recognition}.
\newblock In: \bibinfo{booktitle}{ICCV}. \bibinfo{year}{2015},.
\bibitem[{Savva et~al.(2017)Savva, Yu, Su, Kanezaki, Furuya, Ohbuchi
  et~al.}]{Savva-3DOR2017}
\bibinfo{author}{Savva\xfnm[ M]}, \bibinfo{author}{Yu\xfnm[ F]},
  \bibinfo{author}{Su\xfnm[ H]}, \bibinfo{author}{Kanezaki\xfnm[ A]},
  \bibinfo{author}{Furuya\xfnm[ T]}, \bibinfo{author}{Ohbuchi\xfnm[ R]}, et~al.
\newblock \bibinfo{title}{Shrec'17 track large-scale 3d shape retrieval from
  shapenet core55}.
\newblock In: \bibinfo{booktitle}{Proceedings of the Workshop on 3D Object
  Retrieval}. \bibinfo{year}{2017},.
\bibitem[{{Moscoso Thompson} et~al.(2020){Moscoso Thompson}, Biasotti,
  Giachetti, Tortorici, Werghi, Obeid et~al.}]{Moscoco-SHREC2020}
\bibinfo{author}{{Moscoso Thompson}\xfnm[ E]}, \bibinfo{author}{Biasotti\xfnm[
  S]}, \bibinfo{author}{Giachetti\xfnm[ A]}, \bibinfo{author}{Tortorici\xfnm[
  C]}, \bibinfo{author}{Werghi\xfnm[ N]}, \bibinfo{author}{Obeid\xfnm[ AS]},
  et~al.
\newblock \bibinfo{title}{Shrec 2020: Retrieval of digital surfaces with
  similar geometric reliefs}.
\newblock \bibinfo{journal}{Computers and Graphics} \bibinfo{year}{2020};.
\bibitem[{Wu* et~al.(2022)Wu*, Chen*, Liu, Ren and Wang}]{Wu-NeurIPS2022}
\bibinfo{author}{Wu*\xfnm[ Y]}, \bibinfo{author}{Chen*\xfnm[ Z]},
  \bibinfo{author}{Liu\xfnm[ S]}, \bibinfo{author}{Ren\xfnm[ Z]},
  \bibinfo{author}{Wang\xfnm[ S]}.
\newblock \bibinfo{title}{{CASA}: Category-agnostic skeletal animal
  reconstruction}.
\newblock In: \bibinfo{booktitle}{Neural Information Processing Systems}.
  \bibinfo{year}{2022},.
\bibitem[{Douze et~al.(2021)Douze, Tolias, Pizzi, Papakipos, Chanussot,
  Radenovic et~al.}]{Douze-2021}
\bibinfo{author}{Douze\xfnm[ M]}, \bibinfo{author}{Tolias\xfnm[ G]},
  \bibinfo{author}{Pizzi\xfnm[ E]}, \bibinfo{author}{Papakipos\xfnm[ Z]},
  \bibinfo{author}{Chanussot\xfnm[ L]}, \bibinfo{author}{Radenovic\xfnm[ F]},
  et~al.
\newblock \bibinfo{title}{The 2021 image similarity dataset and challenge}.
\newblock \bibinfo{journal}{arXiv preprint arXiv:210609672}
  \bibinfo{year}{2021};.
\bibitem[{Qi et~al.(2017)Qi, Su, Mo and Guibas}]{Qi-CVPR2017}
\bibinfo{author}{Qi\xfnm[ CR]}, \bibinfo{author}{Su\xfnm[ H]},
  \bibinfo{author}{Mo\xfnm[ K]}, \bibinfo{author}{Guibas\xfnm[ LJ]}.
\newblock \bibinfo{title}{Pointnet: Deep learning on point sets for {3D}
  classification and segmentation}.
\newblock In: \bibinfo{booktitle}{Conference on Computer Vision and Pattern
  Recognition}. \bibinfo{year}{2017}, p. \bibinfo{pages}{652--660}.
\bibitem[{Ma et~al.(2022)Ma, Qin, You, Ran and Fu}]{Ma-2022}
\bibinfo{author}{Ma\xfnm[ X]}, \bibinfo{author}{Qin\xfnm[ C]},
  \bibinfo{author}{You\xfnm[ H]}, \bibinfo{author}{Ran\xfnm[ H]},
  \bibinfo{author}{Fu\xfnm[ Y]}.
\newblock \bibinfo{title}{Rethinking network design and local geometry in point
  cloud: A simple residual mlp framework}.
\newblock \bibinfo{journal}{arXiv preprint arXiv:220207123}
  \bibinfo{year}{2022};.
\bibitem[{Chen et~al.(2020)Chen, Kornblith, Norouzi and Hinton}]{Chen-ICML2020}
\bibinfo{author}{Chen\xfnm[ T]}, \bibinfo{author}{Kornblith\xfnm[ S]},
  \bibinfo{author}{Norouzi\xfnm[ M]}, \bibinfo{author}{Hinton\xfnm[ G]}.
\newblock \bibinfo{title}{A simple framework for contrastive learning of visual
  representations}.
\newblock In: \bibinfo{booktitle}{International conference on machine
  learning}. \bibinfo{year}{2020}, p. \bibinfo{pages}{1597--1607}.
\bibitem[{Tan and Le(2021)}]{Tan-ICML2021}
\bibinfo{author}{Tan\xfnm[ M]}, \bibinfo{author}{Le\xfnm[ Q]}.
\newblock \bibinfo{title}{Efficientnetv2: Smaller models and faster training}.
\newblock In: \bibinfo{booktitle}{International conference on machine
  learning}. \bibinfo{year}{2021}, p. \bibinfo{pages}{10096--10106}.
\bibitem[{Russakovsky et~al.(2015)Russakovsky, Deng, Su, Krause, Satheesh, Ma
  et~al.}]{Russakovsky-IJCV2015}
\bibinfo{author}{Russakovsky\xfnm[ O]}, \bibinfo{author}{Deng\xfnm[ J]},
  \bibinfo{author}{Su\xfnm[ H]}, \bibinfo{author}{Krause\xfnm[ J]},
  \bibinfo{author}{Satheesh\xfnm[ S]}, \bibinfo{author}{Ma\xfnm[ S]}, et~al.
\newblock \bibinfo{title}{{ImageNet Large Scale Visual Recognition Challenge}}.
\newblock \bibinfo{journal}{International Journal of Computer Vision}
  \bibinfo{year}{2015};\bibinfo{volume}{115}(\bibinfo{number}{3}):\bibinfo{pages}{211--252}.
\bibitem[{Vaswani et~al.(2017)Vaswani, Shazeer, Parmar, Uszkoreit, Jones, Gomez
  et~al.}]{Vaswani-NeurIPS2017}
\bibinfo{author}{Vaswani\xfnm[ A]}, \bibinfo{author}{Shazeer\xfnm[ N]},
  \bibinfo{author}{Parmar\xfnm[ N]}, \bibinfo{author}{Uszkoreit\xfnm[ J]},
  \bibinfo{author}{Jones\xfnm[ L]}, \bibinfo{author}{Gomez\xfnm[ AN]}, et~al.
\newblock \bibinfo{title}{Attention is all you need}.
\newblock \bibinfo{journal}{Advances in neural information processing systems}
  \bibinfo{year}{2017};\bibinfo{volume}{30}.
\bibitem[{Hinton et~al.(2012)Hinton, Srivastava, Krizhevsky, Sutskever and
  Salakhutdinov}]{Hinton-2012}
\bibinfo{author}{Hinton\xfnm[ GE]}, \bibinfo{author}{Srivastava\xfnm[ N]},
  \bibinfo{author}{Krizhevsky\xfnm[ A]}, \bibinfo{author}{Sutskever\xfnm[ I]},
  \bibinfo{author}{Salakhutdinov\xfnm[ RR]}.
\newblock \bibinfo{title}{Improving neural networks by preventing co-adaptation
  of feature detectors}.
\newblock \bibinfo{journal}{arXiv preprint arXiv:12070580}
  \bibinfo{year}{2012};.
\bibitem[{Loshchilov and Hutter(2017)}]{Loshchilov-2017}
\bibinfo{author}{Loshchilov\xfnm[ I]}, \bibinfo{author}{Hutter\xfnm[ F]}.
\newblock \bibinfo{title}{Decoupled weight decay regularization}.
\newblock \bibinfo{journal}{arXiv preprint arXiv:171105101}
  \bibinfo{year}{2017};.
\bibitem[{Canny(1986)}]{Canny-TPAMI1986}
\bibinfo{author}{Canny\xfnm[ J]}.
\newblock \bibinfo{title}{A computational approach to edge detection}.
\newblock \bibinfo{journal}{IEEE T-PAMI}
  \bibinfo{year}{1986};(\bibinfo{number}{6}):\bibinfo{pages}{679--698}.
\bibitem[{Madhavan(2021)}]{Madhavan-2021}
\bibinfo{author}{Madhavan\xfnm[ V]}.
\newblock \bibinfo{title}{Artline}.
\newblock
  \bibinfo{howpublished}{\url{https://github.com/vijishmadhavan/ArtLine}};
  \bibinfo{year}{2021}.
\newblock \bibinfo{note}{[Online; accessed 15-March-2023]}.
\bibitem[{Radford et~al.(2021)Radford, Kim, Hallacy, Ramesh, Goh, Agarwal
  et~al.}]{Radford-ICML2021}
\bibinfo{author}{Radford\xfnm[ A]}, \bibinfo{author}{Kim\xfnm[ JW]},
  \bibinfo{author}{Hallacy\xfnm[ C]}, \bibinfo{author}{Ramesh\xfnm[ A]},
  \bibinfo{author}{Goh\xfnm[ G]}, \bibinfo{author}{Agarwal\xfnm[ S]}, et~al.
\newblock \bibinfo{title}{Learning transferable visual models from natural
  language supervision}.
\newblock In: \bibinfo{booktitle}{International conference on machine
  learning}. \bibinfo{year}{2021}, p. \bibinfo{pages}{8748--8763}.
\bibitem[{Dalal and Triggs(2005)}]{Dalal-CVPR2005}
\bibinfo{author}{Dalal\xfnm[ N]}, \bibinfo{author}{Triggs\xfnm[ B]}.
\newblock \bibinfo{title}{Histograms of oriented gradients for human
  detection}.
\newblock In: \bibinfo{booktitle}{CVPR}. \bibinfo{year}{2005}, p.
  \bibinfo{pages}{886--893}.
\bibitem[{Tan and Le(2019)}]{Tan-ICML2019}
\bibinfo{author}{Tan\xfnm[ M]}, \bibinfo{author}{Le\xfnm[ Q]}.
\newblock \bibinfo{title}{Efficientnet: Rethinking model scaling for
  convolutional neural networks}.
\newblock In: \bibinfo{booktitle}{International conference on machine
  learning}. \bibinfo{year}{2019}, p. \bibinfo{pages}{6105--6114}.
\bibitem[{Liu et~al.(2022)Liu, Mao, Wu, Feichtenhofer, Darrell and
  Xie}]{Liu-CVPR2022}
\bibinfo{author}{Liu\xfnm[ Z]}, \bibinfo{author}{Mao\xfnm[ H]},
  \bibinfo{author}{Wu\xfnm[ CY]}, \bibinfo{author}{Feichtenhofer\xfnm[ C]},
  \bibinfo{author}{Darrell\xfnm[ T]}, \bibinfo{author}{Xie\xfnm[ S]}.
\newblock \bibinfo{title}{A convnet for the 2020s}.
\newblock In: \bibinfo{booktitle}{Conference on Computer Vision and Pattern
  Recognition}. \bibinfo{year}{2022}, p. \bibinfo{pages}{11976--11986}.
\bibitem[{Kingma and Ba(2014)}]{Kingma-2014}
\bibinfo{author}{Kingma\xfnm[ DP]}, \bibinfo{author}{Ba\xfnm[ J]}.
\newblock \bibinfo{title}{Adam: A method for stochastic optimization}.
\newblock \bibinfo{journal}{arXiv preprint arXiv:14126980}
  \bibinfo{year}{2014};.
\bibitem[{Nealen et~al.(2005)Nealen, Sorkine, Alexa and
  Cohen-Or}]{Nealen-SIGGRAPH2005}
\bibinfo{author}{Nealen\xfnm[ A]}, \bibinfo{author}{Sorkine\xfnm[ O]},
  \bibinfo{author}{Alexa\xfnm[ M]}, \bibinfo{author}{Cohen-Or\xfnm[ D]}.
\newblock \bibinfo{title}{A sketch-based interface for detail-preserving mesh
  editing}.
\newblock In: \bibinfo{booktitle}{ACM SIGGRAPH}. \bibinfo{year}{2005}, p.
  \bibinfo{pages}{1142--1147}.
\bibitem[{Muzahid et~al.(2020)Muzahid, Wan, Sohel, Wu and
  Hou}]{Muzahid-JAS2020}
\bibinfo{author}{Muzahid\xfnm[ A]}, \bibinfo{author}{Wan\xfnm[ W]},
  \bibinfo{author}{Sohel\xfnm[ F]}, \bibinfo{author}{Wu\xfnm[ L]},
  \bibinfo{author}{Hou\xfnm[ L]}.
\newblock \bibinfo{title}{Curvenet: Curvature-based multitask learning deep
  networks for {3D} object recognition}.
\newblock \bibinfo{journal}{IEEE/CAA Journal of Automatica Sinica}
  \bibinfo{year}{2020};\bibinfo{volume}{8}(\bibinfo{number}{6}):\bibinfo{pages}{1177--1187}.

\end{thebibliography}

\end{document}